\pdfoutput=1

\documentclass[11pt]{article}

\usepackage[final]{acl}

\usepackage{times}
\usepackage{latexsym}
\usepackage{todonotes}
\usepackage{amsmath}
\usepackage{amssymb}

\usepackage[T1]{fontenc}

\usepackage[utf8]{inputenc}

\usepackage{microtype}

\usepackage{inconsolata}

\usepackage{graphicx}

\usepackage{booktabs}
\usepackage{xspace}
\usepackage{multirow}
\usepackage{subcaption} 
\usepackage{relsize}
\usepackage{array}
\usepackage[most]{tcolorbox}
\usepackage{hyperref}

\newcommand{\mmms}{MMMs\xspace}

\newcommand{\metal}{\textsc{M2M}\xspace}

\newcommand{\plusminus}[2]{#1\ensuremath{\pm}\text{\smaller{#2}}}
\newcolumntype{M}[1]{>{\centering\arraybackslash}m{#1}}

\title{\textsc{Multilingual-To-Multimodal} (\metal):\\Unlocking New Languages with Monolingual Text}

\author{Piyush Singh Pasi \\
  Amazon\thanks{Work done outside of Amazon, not related to role directly} \\
  \texttt{piyush.singh.pasi@gmail.com} \\
  }

\begin{document}
\maketitle
\begin{abstract}
Multimodal models excel in English, supported by abundant image–text and audio–text data, but performance drops sharply for other languages due to limited multilingual multimodal resources. Existing solutions rely on machine translation, while advances in multilingual text modeling remain underutilized. We introduce \metal, a lightweight alignment method that learns only a few linear layers—using English text alone—to map multilingual text embeddings into multimodal space. Despite its simplicity, \metal matches baseline performance in English (94.9\% Recall@10) and achieves strong zero-shot transfer (89.5\% Recall@10 averaged across 11 languages, 10 unseen) on XTD Text-to-Image retrieval. Qualitative t-SNE visualizations show that multilingual embeddings align tightly with multimodal representations, while weight analysis reveals that the transformation reshapes embedding geometry rather than performing trivial rotations. Beyond image–text retrieval, \metal demonstrates robustness across datasets and tasks, extending to Audio–Text retrieval and Text-to-Image generation. We release code and checkpoints\footnote{\href{https://github.com/piyushsinghpasi/M2M}{GitHub: piyushsinghpasi/M2M}} along with multilingual evaluation datasets: 
MSCOCO Multilingual 30K\footnote{\href{https://huggingface.co/datasets/piyushsinghpasi/mscoco-multilingual-30k}{HF: piyushsinghpasi/mscoco-multilingual-30k}}, 
AudioCaps Multilingual\footnote{\href{https://huggingface.co/datasets/piyushsinghpasi/audiocaps-multilingual}{HF: piyushsinghpasi/audiocaps-multilingual}}, and 
Clotho Multilingual\footnote{\href{https://huggingface.co/datasets/piyushsinghpasi/clotho-multilingual}{HF: piyushsinghpasi/clotho-multilingual}}.

\end{abstract}
\section{Introduction}

\begin{figure}[ht]
    \centering
    \includegraphics[width=\linewidth]{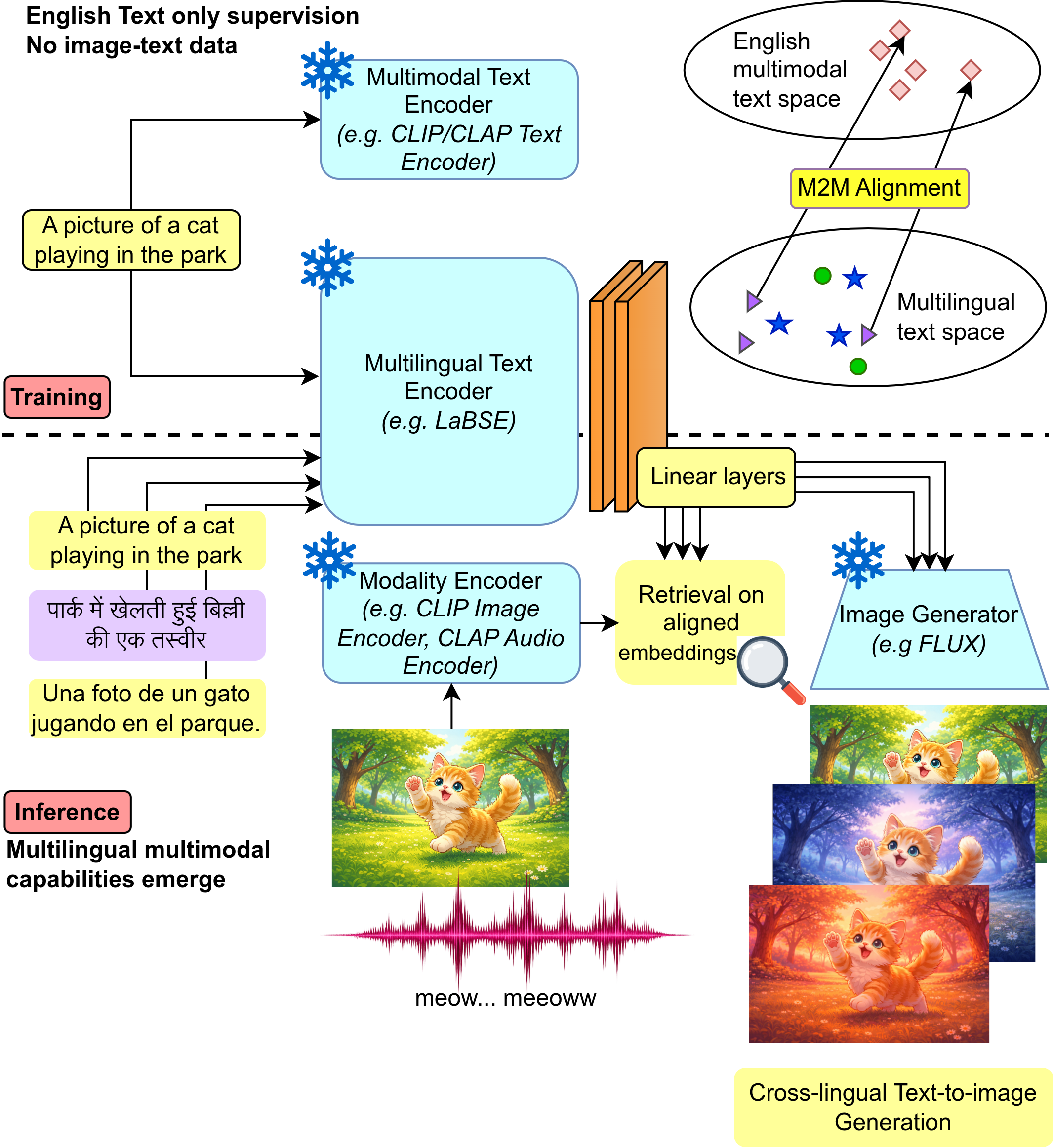}
    \caption{Overview of \metal. Using only English text supervision, we learn a lightweight linear mapping that aligns multilingual text embeddings to a frozen multimodal text space (e.g., CLIP). English acts as a shared anchor during training, aligning multilingual text representations (triangles) to the multimodal text space (diamonds). This alignment implicitly transfers to other languages (stars and circles) without requiring any additional multilingual or multimodal supervision.}
    \label{fig:intro-diagram}
\end{figure}
Humans naturally align information across modalities, associating visual objects/sounds with words. For example, once a person has learned to associate the visual concept of a \emph{cat} with the English word ``cat'', learning that the Spanish word ``gato'' refers to the same concept allows the object–word association for ``gato'' to emerge implicitly, without requiring visual supervision in Spanish.

Existing multimodal models, such as CLIP~\cite{Radford2021LearningTV} and CLAP~\cite{Elizalde2023CLAPLA}, are primarily trained on large-scale English multimodal corpora and rely on explicit multimodal supervision. Extending these models to additional languages typically requires substantial multilingual image–text or audio–text data, either by training from scratch or fine-tuning pretrained models~\cite{carlsson-etal-2022-cross, Yan2024BridgingLG, Koukounas2024jinaclipv2MM}. Acquiring such multilingual multimodal resources is expensive or infeasible, particularly for low-resource languages. By contrast, multilingual text encoders have demonstrated strong cross-lingual generalization using only large-scale text corpora and self-supervised objectives~\cite{devlin-etal-2019-bert, radford2018improving}, but this capability remains largely disconnected from pretrained multimodal representations.

In this work, we propose \metal, a simple, data-efficient, and parameter-efficient approach to bridge multilingual text and multimodal latent spaces using English text as a shared anchor. Inspired by how humans learn, our method does not require explicit multimodal signals for each language—English textual data alone is sufficient for alignment. We learn a lightweight projection map, implemented as a few linear layers, while keeping all pretrained encoders frozen and training with MSE and structure-preserving losses. Despite its simplicity, this alignment enables multilingual text representations to participate directly in multimodal tasks, including retrieval and generation, without observing any multilingual image–text or audio–text pairs. Rather than aiming to replace large-scale multilingual multimodal pretraining, our goal is to show that improved latent-space alignment can recover much of the same capability at a fraction of the data and computational cost.

Prior work has shown the effectiveness of linear projection maps for aligning latent spaces using English multimodal data, mostly in classification settings~\cite{Maiorca2023LatentST, Rosenfeld2022APEAP}. We extend this approach to a broader setting: multilingual and multimodal alignment across retrieval and generative tasks. Our results show that strong multilingual multimodal behavior can emerge from lightweight alignment alone when robust multilingual text encoders are used. To summarize, our contributions are as follows:
\begin{enumerate}
    \item We propose \metal, a lightweight alignment method that maps multilingual text representations into pretrained multimodal latent spaces using only monolingual (English) text data. Despite its simplicity, this approach enables cross-task, cross-modality transfer, allowing multilingual text representations to participate in retrieval and generation tasks without observing any multilingual multimodal data during training.
    \item \metal is highly data-efficient, achieving strong performance with as few as $\sim$1K sentences, and parameter-light, requiring only a few linear layers ($\sim$1–2M parameters). It generalizes across architectures, modalities (image and audio), tasks (Image–Text and Audio–Text retrieval, and Text-to-Image generation), and languages—including those unseen during multimodal pretraining.
    \item We construct synthetic multilingual evaluation benchmarks for multimodal retrieval and generation: (i) Audio–Text retrieval datasets in 33 languages derived from AudioCaps~\cite{kim-etal-2019-audiocaps} (160K samples) and Clotho~\cite{Drossos2019ClothoAA} (172K samples), and (ii) MSCOCO-30K captions translated into 9 additional languages (270K samples) for cross-lingual Text-to-Image generation. These datasets provide a unified and reproducible benchmark for evaluating multilingual multimodal models.
\end{enumerate}

\section{Related Work}
\noindent \textbf{Multilingual Multimodal Models.} Strong multimodal models like CLIP~\cite{Radford2021LearningTV} and CLAP~\cite{Elizalde2023CLAPLA, Wu2022LargeScaleCL} are typically trained on large amounts of English multimodal data (paired image-text and audio-text data). Extending these models to other languages typically requires explicit training on multilingual-multimodal data—either by training from scratch~\cite{Jain2021MURALMM} or by finetuning pretrained models~\cite{Koukounas2024jinaclipv2MM, Yan2024BridgingLG, chen-etal-2023-mclip, ye2024altdiffusion, li-etal-2023-translation}. Some approaches~\cite{carlsson-etal-2022-cross, Chen2022AltCLIPAT, Zhai2021LiTZT} fine-tune only the text encoders while keeping the image encoder frozen, while \citet{aggarwal2020towards} train projection layers on top of frozen encoders using multimodal English data. 

In contrast, our method targets multilingual alignment without relying on multilingual multimodal supervision (e.g., image–text pairs) or encoder fine-tuning. Using simple and intuitive training losses, a small number of linear projection layers, and only English text data, we demonstrate strong multilingual transfer across a broad range of multimodal tasks, including image–text and audio–text retrieval, as well as text-to-image generation.
\\\\
\noindent \textbf{Latent Space Translation.} Latent space translation aims to map representations between distinct latent spaces in order to enable information sharing across modalities or languages. Prior work broadly follows two directions: (i) aligning spaces using relative representations~\cite{Moschella2022RelativeRE, Norelli2022ASIFCD}, and (ii) learning direct transformation mappings between source and target spaces~\cite{gower1975generalized, Maiorca2023LatentST, lahner2024direct}. These techniques have been successfully applied to tasks such as cross-modal classification and generative modeling. A more recent extension is the Inverse Relative Projection method~\cite{maiorca2024latent}, which converts source representations into a relative form before mapping them to a target space, enabling the translation of monolingual text representations into multilingual ones.

Building on this line of work, our approach learns a linear mapping between multilingual and multimodal latent spaces. By leveraging English text as a shared anchor between these spaces, we enable multilingual multimodal transfer without requiring multilingual multimodal training data.
\section{Methodology}
Our method, \metal, is a simple alignment approach that learns a small projection network to align multilingual latent spaces with multimodal latent spaces using English text representations. While we focus on dual-modality multimodal models, the method naturally extends to more than two modalities.  

Consider an English (monolingual) multimodal model $\mathcal{M}_e = (T_e, X_e)$ for language $e$, where $T_e$ is the text encoder and $X_e$ represents any other modality encoder (e.g., image, audio). We assume representations from $T_e$ and $X_e$ are already aligned in a shared latent space using paired multimodal data from language $e$ (e.g., CLIP, CLAP). Let $T_m$ be a multilingual text encoder. For a sentence $s$ in language $e$, let $z_e = T_e(s)$ denote its multimodal representation and $z_m = T_m(s)$ its multilingual representation, with $z_e \in \mathbb{R}^{d_e}$ and $z_m \in \mathbb{R}^{d_m}$. Since both text encoders represent the same sentence $s$, in an \textit{all-aligned} world, $z_e$ and $z_m$ would be identical. In practice, however, they differ due to distinct objectives and training data.  

Our goal\ is therefore to align the multilingual and multimodal latent spaces using English—the common language between these spaces—as an anchor. This alignment enables multimodal tasks on non-English languages, for which no multimodal training data is ever seen; any downstream performance on these languages arises purely from the learned alignment.  

To achieve this, we learn a projection map $\mathcal{F}: \mathbb{R}^{d_m} \rightarrow \mathbb{R}^{d_e}$ that transforms $z_m$ into $z_e$, using text-only alignment data in language $e$ (English). The alignment data must be semantically consistent with the downstream task (e.g., image captions for image–text retrieval, audio captions for audio–text retrieval).  

The projection map $\mathcal{F}$—implemented as a few linear layers—is the only learned component, while all encoders ($T_e$, $T_m$, $X_e$) remain frozen. During inference, multilingual text is encoded by $T_m$, mapped via $\mathcal{F}$, and then directly compared with $X_e$, producing task-compatible representations for retrieval or generation. In this setup, $z_e$ serves as an anchor guiding the translation of multilingual embeddings into the multimodal space. We use mean squared error (MSE) as our primary loss function. 
\begin{align}
z_{m \to e} &= \mathcal{F}(z_m) \\
\mathcal{L}_{\text{align}} &= \text{MSE}(z_{m \to e}, z_e)
\end{align}

\noindent
To derive additional supervision, we enforce structure preservation within each batch $B$. Let $\{z_e^i\}_{i=1}^{|B|}$ and $\{z_{m \to e}^i\}_{i=1}^{|B|}$ denote the target multimodal embeddings and their projected multilingual counterparts. We compute pairwise cosine similarities:
\begin{align}
R_e &= \text{cos\_sim}(z_e^i, z_e^j)_{i,j=1}^{|B|}, \\
R_{m \to e} &= \text{cos\_sim}(z_{m \to e}^i, z_{m \to e}^j)_{i,j=1}^{|B|},
\end{align}
where $R_e, R_{m \to e} \in \mathbb{R}^{|B|\times|B|}$. Let $\text{triu}(\cdot)$ denote the upper-triangular part of a matrix, excluding the diagonal. The structure-preserving loss is then
\begin{align}
\mathcal{L}_{\text{str}} &= \text{MSE}\!\big(\text{triu}(R_e), \text{triu}(R_{m \to e})\big).
\end{align}

\noindent
The final objective combines the alignment and structure-preserving terms:
\begin{align}
\mathcal{L} = \lambda \, \mathcal{L}_{\text{align}} + \beta \, \mathcal{L}_{\text{str}}.
\label{eq:final loss}
\end{align}

\noindent
We experimented with alternative objectives such as L1 loss and similarity loss ($1 - \text{cosine}(z_e, z_{m \to e})$), but these underperform compared to $\mathcal{L}$. MSE is particularly effective because it encourages $z_{m \to e}$ to fully substitute for $z_e$ in the latent space, rather than focusing solely on angular alignment as in contrastive or cosine-based losses. We avoid token- or word-level alignment, which overemphasizes linguistic form over semantics, and do not introduce a reverse mapping $\mathcal{F}_{e \to m}$ since it would disrupt the existing alignment between the other modality encoder $X_e$ and $T_e$. For retrieval tasks, both $\mathcal{L}_{\text{align}}$ and $\mathcal{L}_{\text{str}}$ are computed on L2-normalized embeddings to optimize for downstream evaluation which uses cosine similarity metric. For generative tasks (e.g., text-to-image generation), we omit normalization and $L_{str}$ (see Section~\ref{sec:gen-T2I}).

\section{Exploring Alignment Design Space}
\label{sec:prelim}
We investigate the impact of varying number of linear layers (1, 2, 4), adding or removing residual connections~\cite{He2015DeepRL} in $\mathcal{F}$, and testing different training objectives through ablation studies. 
\begin{table}[t]
\centering
\fontsize{7pt}{9.75pt}\selectfont
\begin{tabular}{l c c c c}\toprule
   Loss&Linear layers&Skip Conn.&T2I&I2T \\ \midrule
    V1: MSE & 2 & Yes & 88.9 & 88.6 \\
    V2: MSE & 2 & No & 89.0 & 88.7 \\
    V3: Similarity Loss & 2 & No & 88.8 & 88.7 \\
    V4: L1 & 2 & No & 86.7 & 83.9 \\
    V5: Ours (eq~\ref{eq:final loss}) & 2 & Yes & 89.2 & \textbf{89.4} \\
    V6: Ours (eq~\ref{eq:final loss}) & 2 & No & \textbf{89.5} & \textbf{89.4} \\
    V7: Ours (eq~\ref{eq:final loss}) & 4 & No & 89.1 & 89.2 \\
    V8: Ours (eq~\ref{eq:final loss}) & 1 & No & 89.3 & 89.3 \\
\bottomrule
\end{tabular}

\caption{Comparison of I2T and T2I Recall@10 (averaged across 11 languages) for different training losses, linear layers, and residual connections (Skip Conn.) with \metal-aligned Jina-CLIP-v1$\times$M-MPNET on XTD test dataset. $\lambda = 48, \beta = 1$.}
\label{tab:arch:losses}
\end{table}

\noindent \textbf{Experimental setup.} We primarily use Jina-CLIP-v1~\cite{Koukounas2024JinaCY} as the multimodal model ($\mathcal{M}_e$) and Multilingual MPNET (M-MPNET)~\cite{reimers-2020-multilingual-sentence-bert} as the multilingual text encoder ($T_m$). Following~\cite{carlsson-etal-2022-cross}, we use a combination of Google Conceptual Captions (GCC)~\cite{Sharma2018ConceptualCA}, MSCOCO~\cite{Lin2014MicrosoftCC}, and VizWiz~\cite{Bigham2010VizWizNR} as our training dataset to learn $\mathcal{F}$. We remove duplicate sentences and create a $N$-sentence training split through random sampling. We experiment with various model architectures and training split sizes (Scaling). Unless specified otherwise, we train for 50 epochs using 250K-sentence training size, batch size of 64, AdamW optimizer~\cite{loshchilov2017decoupled} with a learning rate of 3e-4, weight decay of 1e-2, and a linear learning rate scheduler with 50 warmup steps. All \metal-aligned models are trained on two RTX A5000 24GB Nvidia GPUs. For validation, we use XTD~\cite{aggarwal2020towards} English image-text pairs, saving the best checkpoint based on the mean of Text-to-Image (T2I) and Image-to-Text (I2T) recall. Both T2I and I2T recalls are averaged across Recall@1,5,10. We evaluate these experiments on Image-to-Text retrieval task using the XTD test dataset.\\

\begin{figure}[t]
    \centering
    \includegraphics[width=\linewidth]{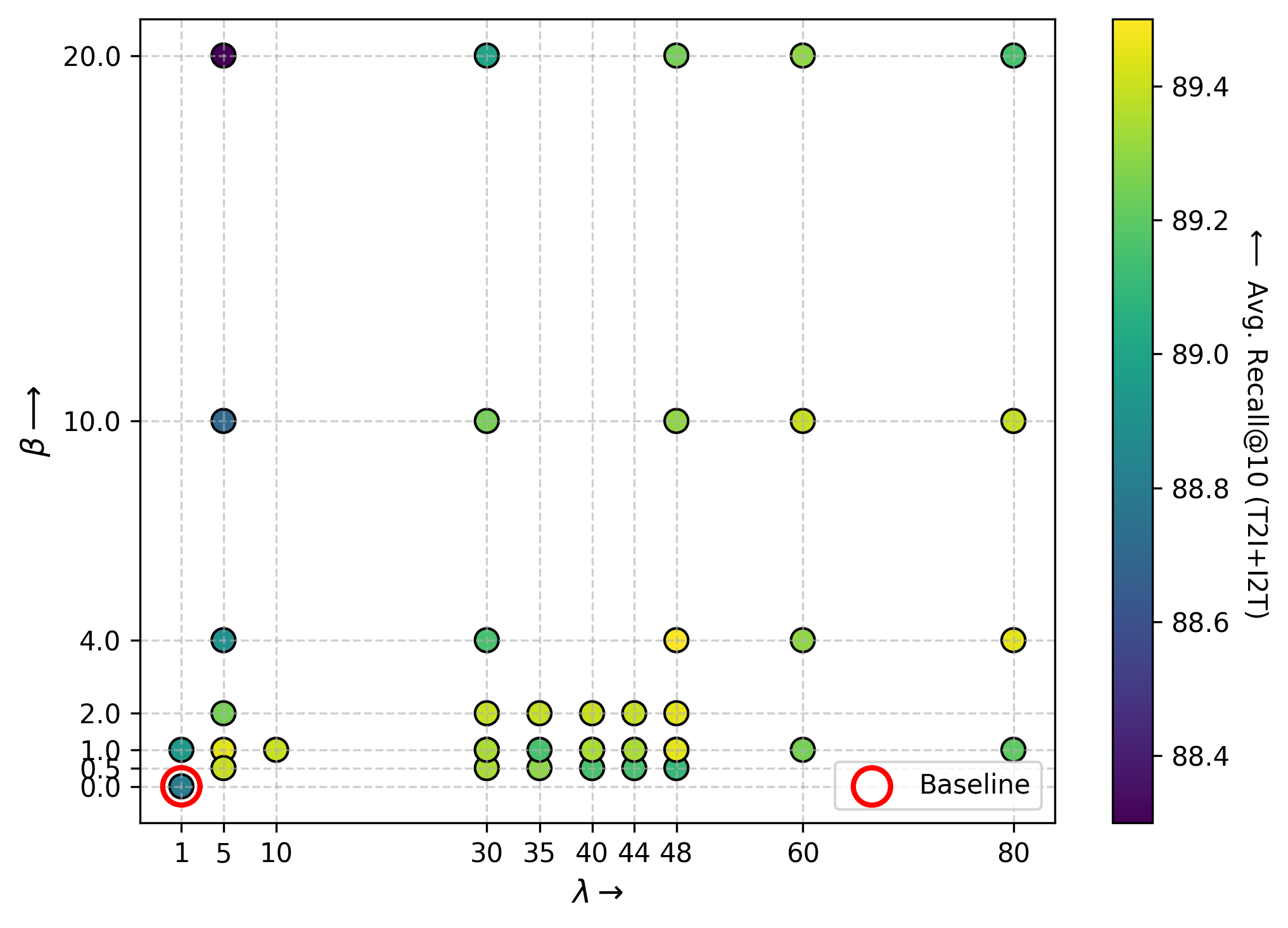}
    \caption{Impact of $\lambda$ and $\beta$ on XTD image–text retrieval. Increasing $\lambda$ while reducing $\beta$ leads to consistent performance gains.}
    \label{fig:hyp-lambda-beta}
\end{figure}

\noindent \textbf{Results and discussion.} Table~\ref{tab:arch:losses} shows that our proposed loss (eq.~\ref{eq:final loss}) consistently outperforms alternatives. Two linear layers and no skip connection (row V6) achieves the best performance, yielding absolute gains of up to 0.7\% over MSE (row V2) and similarity loss (row V3), and around 3--5\% over L1 loss (row V4). Varying linear layers or using a skip connection has minor effects (rows V6-V8), indicating the model is robust to these architectural choices. Assigning higher weight to $\mathcal{L}_{\text{align}}$ ($\lambda=48$) versus $\mathcal{L}_{\text{str}}$ ($\beta=1$) yields a 0.5\% gain compared to equal weighting ($\lambda=1,\beta=1$), as shown in Figure~\ref{fig:hyp-lambda-beta}. Overall, the combination of two linear layers ($\sim$1M parameters) without a skip connection and $\lambda=48, \beta=1$ provides the strongest results. See appendix~\ref{app:prelim} for detailed numbers.

\begin{figure}[ht]
    \centering
    \includegraphics[width=\linewidth]{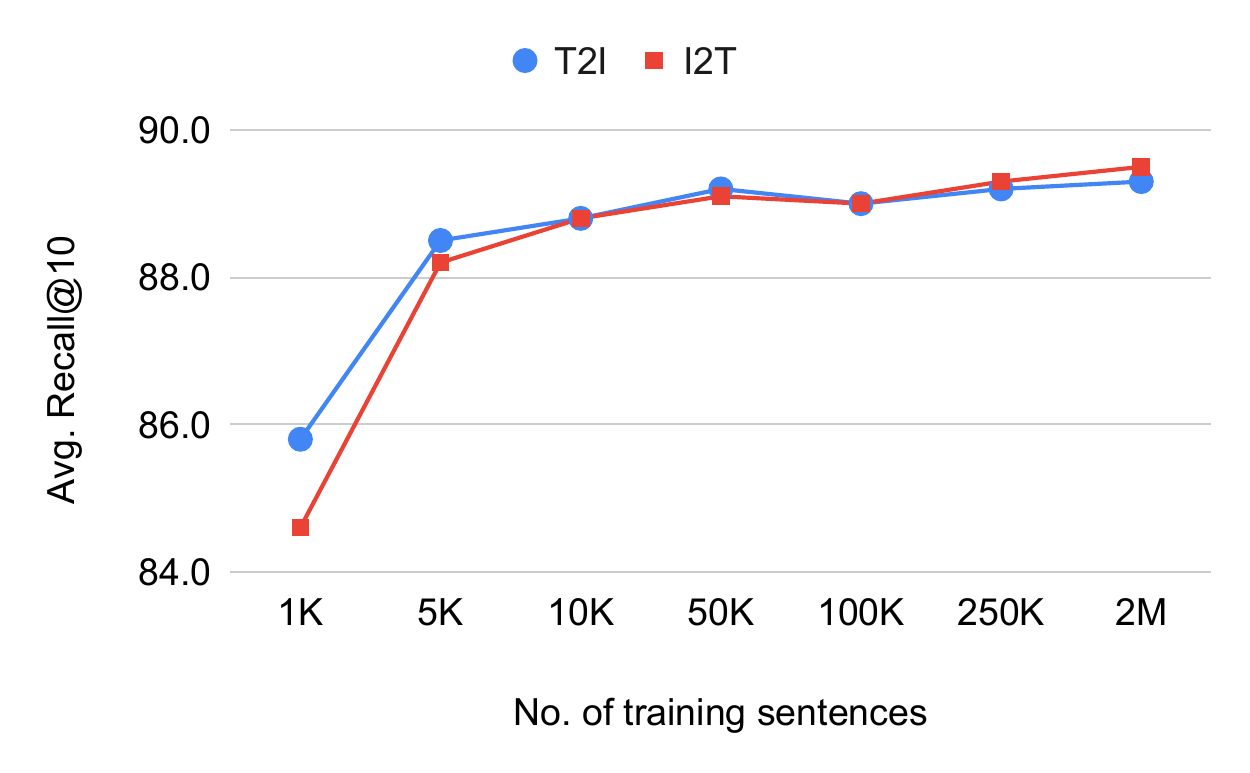}
    \caption{Effect of scaling training data on XTD eval set for \metal-aligned model Jina-CLIP-v1 $\times$ M-MPNET. Recall@10 averaged across all languages in XTD.}
    \label{fig:scale}
\end{figure}

Data scaling experiments (Figure~\ref{fig:scale}) using the optimal configuration (2 linear layers, no residuals, $\lambda=48, \beta=1$) show that \metal achieves 85.8\% Avg. Recall@10 with just 1,000 English sentences, without any multilingual or multimodal data. Performance saturates beyond 250K sentences; scaling to 2M sentences provides minimal improvements (0.1--0.2\%).  

\begin{figure}[ht]
    \centering
    \includegraphics[width=\linewidth]{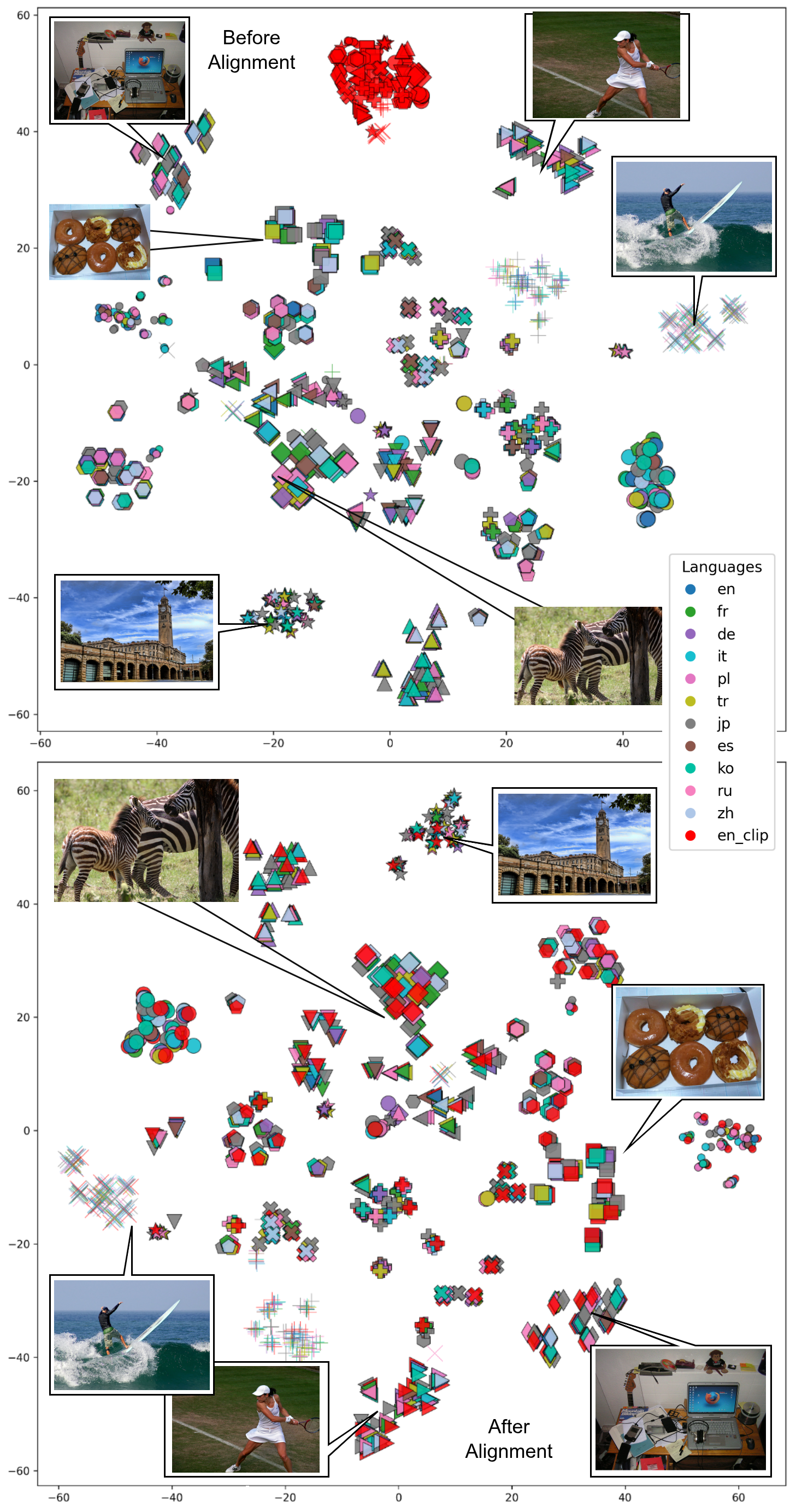}
    \caption{t-SNE visualization (perplexity = 32) of text embeddings before and after alignment. Marker shapes denote visual clusters and colors indicate languages, with English Jina-CLIP-v1 text embeddings ($z_e$ or en\_clip) in red. Before alignment (top), text embeddings ($z_m$ \& $z_e$) are fragmented; after alignment (bottom), multilingual captions ($z_{m \to e}$) and Jina-CLIP-v1 text embeddings ($z_e$) align closely with shared visual clusters.
    }
    \label{fig:tsne}
\end{figure}

\noindent \textbf{Alignment quality.} For the best configuration (row V6 in Table~\ref{tab:arch:losses}), we visualize alignment using t-SNE (Figure~\ref{fig:tsne}). To avoid language bias, we first cluster image embeddings of Jina-CLIP-v1 (J-CLIP) from the XTD test set using KMeans (K=100), then select 17 clusters via farthest-cluster sampling from the 50 largest clusters, excluding very small clusters. From each cluster, we sample up to 10 points to prevent overcrowding. We compute t-SNE jointly for text embeddings from J-CLIP ($z_e$) and M-MPNET ($z_m$) before alignment (Figure~\ref{fig:tsne}, top), and for J-CLIP embeddings ($z_e$) with aligned embeddings ($z_{m \to e}$) after alignment (Figure~\ref{fig:tsne}, bottom) to visualize the effect of \metal. As shown, J-CLIP and M-MPNET embeddings occupy distinct regions before alignment; after alignment, J-CLIP embeddings align with the multilingual embeddings across clusters, demonstrating effective cross-lingual alignment.\\

\noindent \textbf{Weight analysis.} Using our best configuration (row V6 in Table~\ref{tab:arch:losses}), we analyze the effective linear map $W_{eff} = W_2 W_1$. The singular value spectrum revealed that the map focuses on a compact, semantically relevant subspace, with an effective rank of $\sim 204$. The orthogonality deviation ($\|W^\top W - I\|_F \approx 554$) indicates that the transformation involves substantial \emph{mixing and rescaling} rather than a simple rotation. The effective bias is small ($\|\mathbf{b}_{\text{eff}}\| \approx 1.3$), suggesting that alignment primarily reshapes the geometry of embeddings rather than just shifting them. Because the effective rank is lower than $d_e=d_m=768$, Jina-CLIP-v1 text embeddings likely capture language-specific features that are absent in the language-agnostic M-MPNET space. Pairwise cosine distances between nearly identical sentences further support this hypothesis: Jina-CLIP-v1 embeddings vary the most (0.03–0.08), multilingual MPNET embeddings are tighter (0.01–0.04), and the mapped embeddings fall in between (0.01–0.05). This confirms that Jina-CLIP-v1 embeddings encode language-specific style variations, whereas the mapped embeddings preserve semantic consistency across sentence variants. See additional details and box plots in Appendix~\ref{app:weight analysis}.
\begin{table*}
\centering
\fontsize{7pt}{9.75pt}\selectfont
\begin{tabular}{p{3.4cm} p{0.25cm} p{0.25cm} p{0.25cm} p{0.25cm} p{0.25cm} p{0.25cm} p{0.25cm} p{0.25cm} p{0.25cm} p{0.25cm} p{0.25cm} p{0.25cm} p{0.25cm} p{0.25cm} p{0.25cm} p{0.25cm} p{0.25cm} p{0.25cm}}
\toprule
\multirow{2}{*}{Models} & \multicolumn{12}{c}{XTD-T2I} & \multicolumn{2}{c}{XTD-I2T} & \multicolumn{2}{c}{XM3600} & \multicolumn{2}{c}{Multi30K} \\ 
 & Avg. & de & en & es & fr & it & jp & ko & pl & ru & tr & zh & Avg. & en & T2I & I2T & T2I & I2T \\ \midrule
\multicolumn{19}{l}{\textbf{English-only Vision-Language Models}\vspace{0.3em}}\\ 
E1: CLIP (ViT-L 336px) & 35.7 & 55.4 & 92.5 & 64.1 & 67.0 & 53.7 & 18.7 & 2.7 & 15.6 & 5.0 & 13.2 & 4.9 & 43.2 & 94.1 & 14.0 & 23.7 & 54.9 & 63.7 \\
E2: Jina-CLIP-v1 & 37.4 & 61.5 & \textbf{95.0} & \textbf{67.8} & 77.4 & 58.3 & 9.8 & 1.9 & 16.8 & 4.4 & 10.9 & \textbf{7.5} & 39.5 & \textbf{95.8} & 20.3 & 26.5 & 58.9 & 59.6 \\
E3: K-ALIGN & \textbf{47.6} & \textbf{73.3} & 94.0 & 67.1 & \textbf{80.0} & \textbf{72.8} & \textbf{26.2} & \textbf{12.6} & \textbf{37.6} & \textbf{34.0} & \textbf{19.1} & 7.0 & \textbf{53.1} & 93.8 & \textbf{22.9} & \textbf{31.0} & \textbf{67.5} & \textbf{70.1} \\ \midrule 

\multicolumn{19}{l}{\textbf{Multilingual Vision-Language Models Trained on Supervised Multimodal and/or Multilingual Data}\vspace{0.3em}}  \\
T1: mUSEM3L & 74.9 & 73.5 & 85.3 & 76.7 & 78.9 & 78.9 & 67.8 & 70.7 & 71.7 & 73.6 & 70.9 & 76.1 & {\hfill\textendash} & {\hfill\textendash} & {\hfill\textendash} & {\hfill\textendash} & {\hfill\textendash} & {\hfill\textendash} \\ 

T2: MCLIP-ST & 76.4 & 78.7 & 88.5 & 78.2 & 79.8 & 79.3 & 68.6 & 63.1 & 75.6 & 74.7 & 74.4 & 79.4 & 78.6 & 90.4 & 48.7 & 60.6 & 80.7 & 83.4 \\

T3: ALIGN-Base & 82.2* & {\hfill\textendash} & {\hfill\textendash} & 88.8 & {\hfill\textendash} & 87.9 & {\hfill\textendash} & 76.6 & 79.8 & 82.3 & 73.5 & 86.5 & {\hfill\textendash} & {\hfill\textendash} & {\hfill\textendash} & {\hfill\textendash} & {\hfill\textendash} & {\hfill\textendash} \\
T4: MURAL-Large & 90.2* & {\hfill\textendash} & {\hfill\textendash} & 92.9 & {\hfill\textendash} & 91.8 & {\hfill\textendash} & 88.1 & 91.0 & 87.2 & 89.5 & 89.7 & {\hfill\textendash} & {\hfill\textendash} & {\hfill\textendash} & {\hfill\textendash} & {\hfill\textendash} & {\hfill\textendash} \\
T5: LABSE ViT-L/14 & 87.2 & 89.6 & 91.6 & 89.5 & 89.9 & 90.1 & 73.9 & 80.8 & 89.8 & 85.5 & 89.8 & 88.9 & 90.8 & 94.9 & 73.2 & 83.6 & 90.9 & 93.7 \\

T6: XLM-R-L ViT-B/32 & 88.0 & 88.7 & 91.8 & 89.1 & 89.4 & 89.8 & 81.0 & 82.1 & 91.4 & 86.1 & 88.8 & 89.3 & 89.9 & 91.7 & 75.2 & 84.5 & 89.2 & 91.0 \\

T7: XLM-R ViT-L/14 & 89.0 & 90.6 & 92.4 & 91.0 & 90.0 & 91.1 & 81.9 & 85.2 & 91.3 & 85.8 & 90.3 & 89.7 & 92.2 & 94.5 & 76.4 & 85.0 & 92.2 & 94.4 \\

T8: XLM-R-L ViT-B/16+ & 92.0 & 93.0 & 95.0 & 93.6 & 93.1 & 93.1 & 84.2 & 89.0 & 94.4 & 90.0 & 93.0 & 94.0 & 93.2 & 96.1 & \textbf{81.8} & \textbf{87.1} & 93.9 & 94.2 \\

T9: Jina-CLIP-v2 & {\textbf{92.6}} & 92.5 & 92.8 & 88.9 & \textbf{95.5} & 93.2 & \textbf{94.1} & 90.6 & 94.9 & 90.7 & \textbf{93.5} & 91.4 & 93.2 & 92.7 & 81.1 & 85.7 & 93.8 & 94.0 \\

T10: $\text{AltCLIP}_{M9}$ & \textbf{93.7}* & {\hfill\textendash} & 95.4 & 94.1	& 92.9	& 94.2	& 91.7	& \textbf{94.4}	& {\hfill\textendash}	& 91.8	& {\hfill\textendash}	& \textbf{95.1} & {\hfill\textendash} & {\hfill\textendash} & {\hfill\textendash} & {\hfill\textendash} & {\hfill\textendash} & {\hfill\textendash} \\

T11: SigLIP  & 67.2 & 87.9 & 96.7 & 93.3 & 91.0 & 90.2 & 19.7 & 25.8 & 75.5 & 69.3 & 63.3 & 26.2 & 71.6 & \textbf{98.3} & 40.1 & 51.9 & 85.4 & 87.7 \\

T12: SigLIP2 & 92.6 & \textbf{94.6} & 96.7 & \textbf{95.8} & 95.0 & \textbf{96.1} & 80.2 & 91.3 & \textbf{95.8} & \textbf{92.1} & 91.7 & 89.1 & 93.7 & 97.9 & 74.6 & 81.5 & \textbf{96.2} & 96.2 \\

T13: LLM2CLIP & 92.1 & 92.0 & \textbf{97.3} & 93.8 & 92.1 & 93.6 & 91.9 & 88.6 & 93.2 & 89.4 & 87.7 & 93.8 & \textbf{94.0} & 97.8 & 72.8 & 83.5 & 96.0 & \textbf{96.7} \\

\midrule

\multicolumn{19}{l}{\textbf{\metal-aligned Multilingual Multimodal models using English-only Text data}\vspace{0.3em}} \\
M1: Jina-CLIP-v1 $\times$ LaBSE & 82.7 & 82.4 & 86.5 & 83.6 & 84.8 & 85.0 & 76.5 & 80.3 & 85.4 & 80.7 & 81.4 & 82.6 & 80.0 & 86.8 & 62.9 & 65.6 & 79.0 & 75.7 \\
M2: Jina-CLIP-v1 $\times$ M-MiniLM & 86.4 & 86.7 & 93.8 & 88.1 & 88.4 & 87.6 & 80.5 & 74.8 & 89.0 & 85.2 & 86.3 & \textbf{90.2} & 84.9 & 93.7 & 57.7 & 64.5 & 88.0 & 85.9 \\  
M3: Jina-CLIP-v1 $\times$ JinaTextV3 & 88.0 & \textbf{91.1} & \textbf{94.9} & 89.6 & \textbf{90.6} & 90.9 & 80.2 & 80.9 & 90.4 & 85.5 & 88.0 & 85.9 & 87.5 & 95.0 & \textbf{67.2} & 72.2 & 87.8 & 87.5 \\
M4: Jina-CLIP-v1 $\times$ M-MPNET & \textbf{89.5} & 90.5 & 94.7 & \textbf{91.9} & 90.5 & \textbf{91.1} & \textbf{82.4} & \textbf{85.8} & \textbf{91.2} & \textbf{86.8} & \textbf{89.2} & 89.9 & \textbf{89.4} & \textbf{95.2} & 66.3 & \textbf{73.1} & 89.9 & 89.7 \\
M5: CLIP $\times$ M-MPNET & 84.6 & 85.6 & 90.8 & 86.6 & 85.3 & 86.2 & 79.1 & 80.5 & 85.1 & 82.3 & 84.6 & 84.5 & 86.2 & 93.9 & 56.2 & 67.3 & 90.1 & \textbf{92.1} \\
M6: K-ALIGN $\times$ M-MPNET & 86.8 & 87.5 & 92.6 & 89.6 & 87.9 & 87.8 & 78.9 & 83.0 & 89.0 & 83.5 & 87.0 & 87.9 & 86.0 & 94.4 & 59.0 & 68.2 & 91.0 & 90.2 \\ 
M7: SigLIP $\times$ M-MPNET	& 86.1 & 87.2 & 92.3 & 86.7 & 87.2 & 87.3 & 80.9 & 82.8 & 87.0 & 83.0 & 85.6 & 87.0 & 85.9 & 94.5 & 59.1 & 65.3 & \textbf{93.5} & 91.4 \\

\bottomrule

\end{tabular}

\caption{Comparison of \metal-aligned model performance with English and Multilingual CLIP-like models using Recall@10 across datasets. Results include reported XTD-T2I numbers for T1, T3-T8, T10 and rest are computed using available checkpoints. * denotes average is computed over only supported languages.}
\label{tab:main}

\end{table*}

\section{Image-Text Retrieval}
\label{sec:image-text}

\noindent \textbf{Experimental setup.} 
We evaluate English vision–language models ($\mathcal{M}_e$): CLIP~\cite{Radford2021LearningTV}, Jina-CLIP-v1~\cite{Koukounas2024JinaCY}, K-ALIGN~\cite{kakaobrain2022coyo-align}, \& SigLIP\footnote{\href{https://huggingface.co/google/siglip-so400m-patch14-384}{HF: google/siglip-so400m-patch14-384}}~\cite{zhai2023sigmoid}, and multilingual text encoders ($T_m$): LaBSE~\cite{Feng2020LanguageagnosticBS}, M-MPNET and M-MiniLM~\cite{reimers-2020-multilingual-sentence-bert}, and Jina-Text-v3~\cite{sturua2024jina}. 
Aligned models are denoted as $\mathcal{M}_e \times T_m$ (e.g., CLIP $\times$ LaBSE).  

We compare \metal-aligned models against:  
(i) English-only $\mathcal{M}_e$ (CLIP, Jina-CLIP-v1, K-ALIGN), and  
(ii) multilingual multimodal baselines (\mmms): mUSEM3L~\cite{aggarwal2020towards}, MCLIP-ST (Multilingual CLIP~\cite{reimers-2020-multilingual-sentence-bert} from SentenceTransformers\footnote{\url{https://www.sbert.net/}}), MURAL-Large~\cite{Jain2021MURALMM}, ALIGN-Base~\cite{jia2021scaling} (via MURAL), XLM-R ViT variants (ViT-L/14, ViT-B/32, ViT-B/16+) \& LaBSE ViT-L/14~\cite{carlsson-etal-2022-cross}, Jina-CLIP-v2~\cite{Koukounas2024jinaclipv2MM}, $\text{AltCLIP}_{M9}$~\cite{Chen2022AltCLIPAT}, SigLIP, SigLIP2\footnote{\href{https://huggingface.co/google/siglip2-so400m-patch14-384}{HF: google/siglip2-so400m-patch14-384}}~\cite{tschannen2025siglip}, and LLM2CLIP\footnote{\href{https://huggingface.co/microsoft/LLM2CLIP-Openai-L-14-336}{HF: microsoft/LLM2CLIP-Openai-L-14-336}}~\cite{huang2024llm2clip}. 
Supported languages are listed in Appendix~\ref{app:list-lang}.

\noindent \textbf{Evaluation.} We evaluate on three multilingual image-text datasets: XTD (11 languages)~\cite{aggarwal2020towards}, which includes MIC~\cite{rajendran-etal-2016-bridge} (de, fr) and STAIR Captions~\cite{yoshikawa-etal-2017-stair} (jp); XM3600 (36 languages)~\cite{thapliyal-etal-2022-crossmodal}; and Multi30K (4 languages)~\cite{W16-3210, elliott-EtAl:2017:WMT, barrault2018findings}. Following prior work~\cite{aggarwal2020towards, Jain2021MURALMM, carlsson-etal-2022-cross}, we use Recall@10 with cosine similarity as the ranking score. For XTD, we report Text-to-Image retrieval scores for all languages and the average Recall@10. For XM3600, Multi30K, and Image-to-Text retrieval tasks, we report only the mean Recall@10 across all languages; per-language results are provided in Appendix~\ref{app:image-text}.

\noindent \textbf{Results \& Analysis.}  
For the XTD Text-to-Image (T2I) task, the \metal-aligned $\text{Jina-CLIP-v1}\times\text{M-MPNET}$ model (row M4, Table~\ref{tab:main}) outperforms several multilingual multimodal models trained with multilingual and/or multimodal paired data (rows T1–T3, T5–T7, T11). On English, our aligned models (rows M3 and M4) match the performance of English-trained baselines (rows E1–E3), indicating that alignment does not degrade monolingual performance. Importantly, these gains arise solely from representation alignment, as no new multilingual or multimodal supervision is introduced during training. For subsequent comparisons, we adopt Jina-CLIP-v2 as the state-of-the-art reference model, since it achieves the strongest average performance across different languages and datasets.

On XTD, our best \metal-aligned model (row M4) performs 3.1\% lower on T2I and 3.8\% lower on Image-to-Text (I2T) compared to SOTA. This gap is expected, as models like Jina-CLIP-v2 are explicitly trained on massive multilingual-multimodal data—$\sim$400M non-English image-text pairs from CommonPool~\cite{Gadre2023DataCompIS} and 1.2M multilingual synthetic captions. For Multi30K, the performance gap is similar: 3.9\% for T2I and 4.3\% for I2T. For XM3600, the gap widens to 14.8\% (T2I) and 12.6\% (I2T), likely due to the larger retrieval space (Multi30K and XTD test sets have 1K instances, while XM3600 has 3,600 images and $\sim$7K captions). Detailed per-language results for XM3600 and Multi30K are provided in Appendix~\ref{app:image-text}.


\section{Audio-Text Retrieval}
\label{sec:clap}

\begin{table}[!htpb]
\centering
\fontsize{7pt}{9.75pt}\selectfont
\begin{tabular}{l c c c c}
\toprule
\multirow{2}{*}{Models} & \multicolumn{2}{c}{AudioCaps} & \multicolumn{2}{c}{Clotho} \\
 & Avg. & en & Avg. & en\\ 
 \midrule
 \multicolumn{5}{l}{\textbf{English-only LAION-CLAP Models}\vspace{0.3em}}\\ 
CF: HTSAT-Fused &  {\textendash} & 70.3/82.5* &  {\textendash} & \textbf{49.9}/\textbf{55.4}* \\
CG: General &  {\textendash} & \textbf{83.4} & {\textendash} & 49.3 \\ \midrule
\multicolumn{5}{l}{\textbf{\metal-aligned Multilingual CLAP models}\vspace{0.3em}} \\

CM1: CF $\times$ M-MPNET & 46.7 & 62.7 & 36.7 & 46.7 \\
CM2: CG $\times$ M-MPNET & \textbf{54.2} & \textbf{77.4} & \textbf{36.8} & \textbf{47.6} \\
\bottomrule
\end{tabular}
\caption{Performance comparison of Audio-Text Models on AudioCaps and Clotho datasets using Recall@10 for Text-to-Audio (T2A) retrieval, averaged across M-MPNET supported languages. * denotes reported numbers from \citet{Wu2022LargeScaleCL} and rest are computed from checkpoints.}
\label{tab:clap}
\end{table}

\noindent \textbf{Experimental Setup.} We use LAION-CLAP~\cite{Wu2022LargeScaleCL} as the audio-text multimodal model ($\mathcal{M}_e$) and align it with M-MPNET ($T_m$). We experiment with two variants of LAION-CLAP:  
(i) CLAP-HTSAT-fused, trained on AudioCaps~\cite{kim-etal-2019-audiocaps}, Clotho~\cite{Drossos2019ClothoAA}, and LAION-Audio-630k~\cite{Wu2022LargeScaleCL}; and  
(ii) CLAP-General, trained on additional speech and music data. For alignment, we use English captions from AudioCaps, Clotho, and WavCaps~\cite{Mei2023WavCapsAC}. The AudioCaps validation set is used to select the best checkpoint.\\

\noindent \textbf{Synthetic Evaluation Datasets.}  
Due to the lack of multilingual audio–text evaluation datasets, we extend the AudioCaps and Clotho test sets to 33 additional languages using machine translation. In particular, we translate 4,875 AudioCaps captions (although evaluation was performed on only 4,785 captions because 18 audio files were unavailable during dataset creation) and 5,225 Clotho captions. For 11 Indic languages\footnote{bn, gu, hi, kn, ml, mr, ne, pa, ta, te, ur}, we use the English-to-Indic translation model from IndicTrans2~\cite{Gala2023IndicTrans2TH}. For the remaining 22 languages\footnote{ar, zh-Hans, zh-Hant, cs, nl, fr, de, el, he, id, it, ja, ko, fa, pl, pt, ro, ru, es, tr, uk, vi}, we use Aya-23-35B model~\cite{Aryabumi2024Aya2O}.

Based on Aya-23-35B’s reported results on the FLoRes-200 test set~\cite{costa2022no} and manual spot checks, we assume that the translations for the 22 languages are of reasonably high quality. The FLoRes-200 test set is also used to identify the optimal prompt for translation. To evaluate translation quality for Indic languages, we back-translate to English using IndicTrans2 (Indic-to-English). Across the 11 Indic languages, AudioCaps achieves a mean spBLEU~\cite{post-2018-call} of 48.7 and chrF++~\cite{popovic-2017-chrf} of 63.6, while Clotho achieves 47.4 and 59.6, respectively.  

Additional details on dataset licenses and translation quality assessment are provided in Appendix~\ref{app:license} and~\ref{app:trans}. Due to the lack of comparable multilingual baselines, we report Recall@10 for our method only on these synthetic multilingual test sets. Language-wise Recall@10 scores are provided in Appendix~\ref{app:clap} for both AudioCaps and Clotho. \\

\noindent \textbf{Results \& Analysis.}  
Table~\ref{tab:clap} shows that our method generalizes effectively to modalities beyond images. On AudioCaps, our approach performs 6\% below the SOTA on Text-to-Audio retrieval (T2A), while on Clotho the gap is 2.3\% (T2A). To understand this gap, we compute Text-to-Text (T2T) Recall@10 on XM3600 (image-text) and AudioCaps (audio-text), leveraging multiple captions per instance. M-MPNET achieves 62.1\% T2T Recall@10 on XM3600, comparable to Jina-CLIP-v1 (63.8\%), but only 73.8\% on AudioCaps, substantially lower than CLAP-General (80.2\%). This suggests that M-MPNET, while effective for image-caption encoding, underperforms for audio-caption encoding.  

\begin{figure*}[t] 
    \centering
    \begin{subfigure}{0.157\textwidth}
        \includegraphics[width=\linewidth]{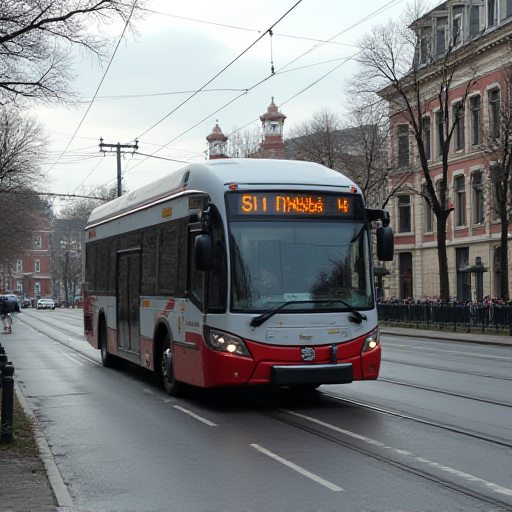}
        \caption{FLUX (en)}
    \end{subfigure}
    \hspace*{\fill}
    \begin{subfigure}{0.157\textwidth}
        \includegraphics[width=\linewidth]{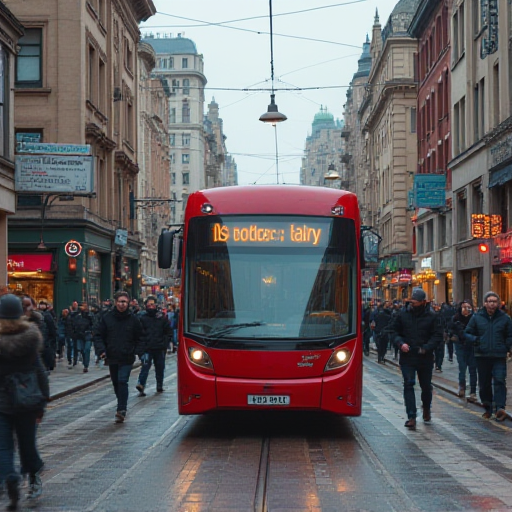}
        \caption{FLUX CLIP (en)}
    \end{subfigure}
    \hspace*{\fill}
    \begin{subfigure}{0.157\textwidth}
        \includegraphics[width=\linewidth]{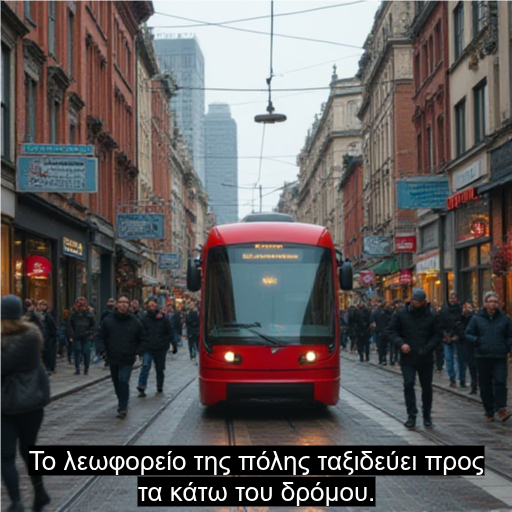}
        \caption{Ours (el)}
    \end{subfigure}
    \hspace*{\fill}
    \begin{subfigure}{0.157\textwidth}
        \includegraphics[width=\linewidth]{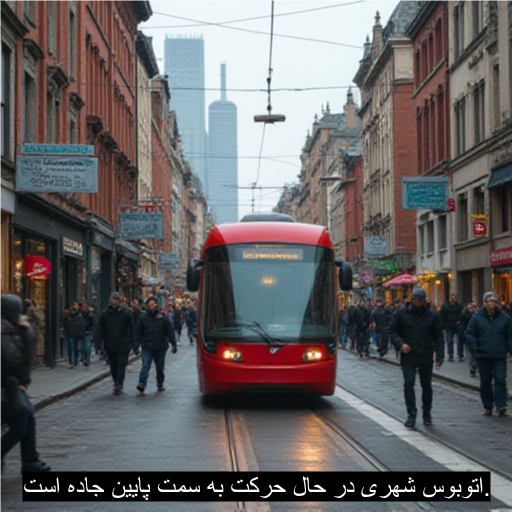}
        \caption{Ours (fa)}
    \end{subfigure}
    \hspace*{\fill}
    \begin{subfigure}{0.157\textwidth}
        \includegraphics[width=\linewidth]{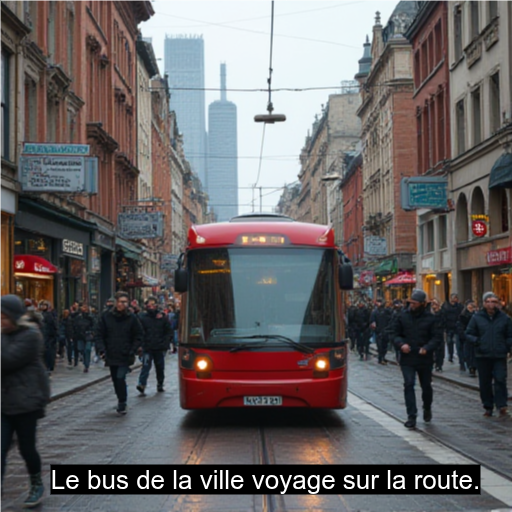}
        \caption{Ours (fr)}
    \end{subfigure}
    \hspace*{\fill}
    \begin{subfigure}{0.157\textwidth}
        \includegraphics[width=\linewidth]{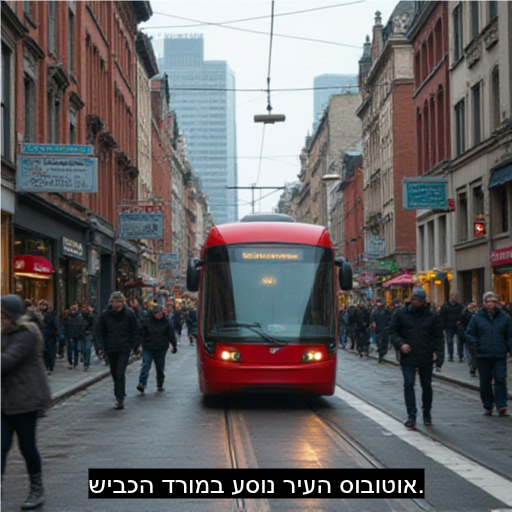}
        \caption{Ours (he)}
    \end{subfigure}

    \vspace{10pt} 

    \begin{subfigure}{0.157\textwidth}
        \includegraphics[width=\linewidth]{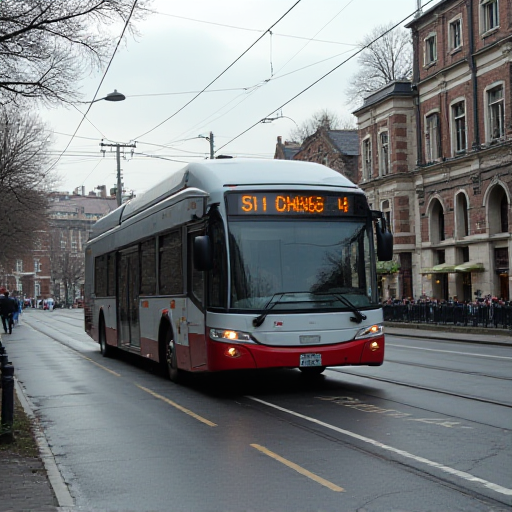}
        \caption{FLUX-T5 (en)}
    \end{subfigure}
    \hspace*{\fill}
    \begin{subfigure}{0.157\textwidth}
        \includegraphics[width=\linewidth]{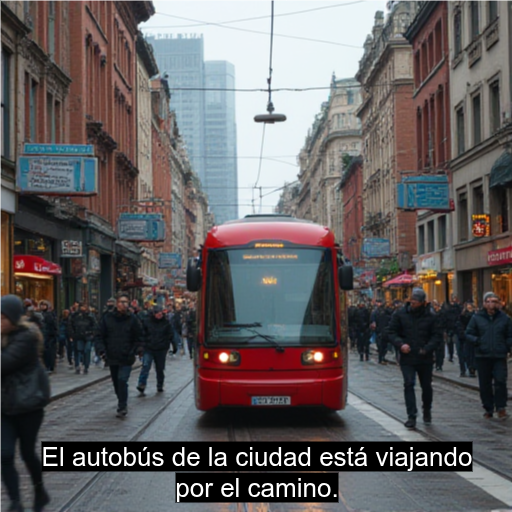}
        \caption{Ours (es)}
    \end{subfigure}
    \hspace*{\fill}
    \begin{subfigure}{0.157\textwidth}
        \includegraphics[width=\linewidth]{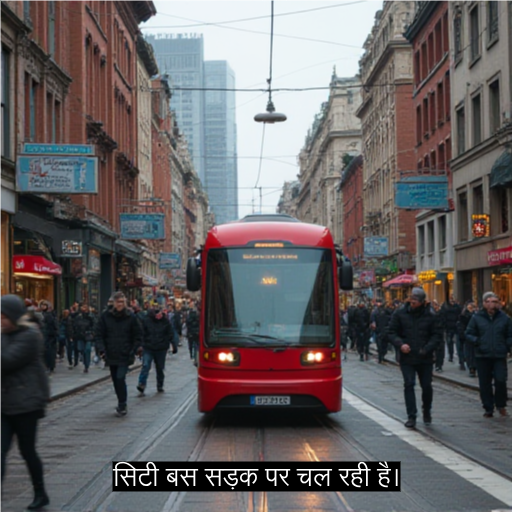}
        \caption{Ours (hi)}
    \end{subfigure}
    \hspace*{\fill}
    \begin{subfigure}{0.157\textwidth}
        \includegraphics[width=\linewidth]{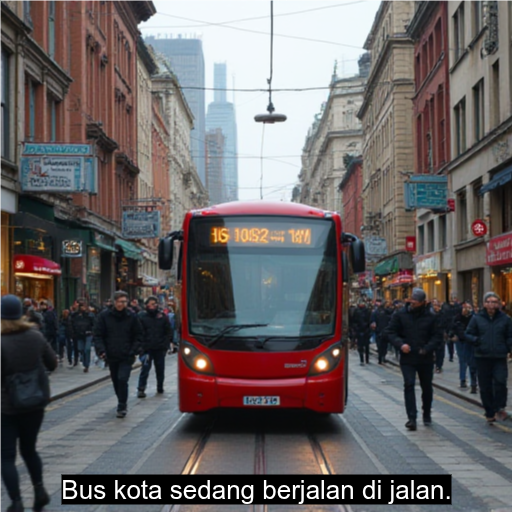}
        \caption{Ours (id)}
    \end{subfigure}
    \hspace*{\fill}
    \begin{subfigure}{0.157\textwidth}
        \includegraphics[width=\linewidth]{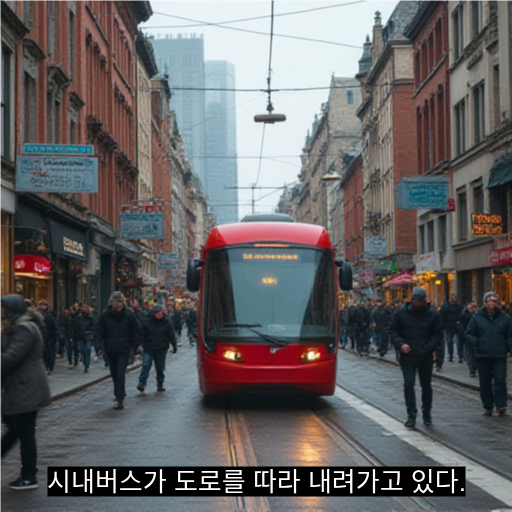}
        \caption{Ours (ko)}
    \end{subfigure}
    \hspace*{\fill}
    \begin{subfigure}{0.157\textwidth}
        \includegraphics[width=\linewidth]{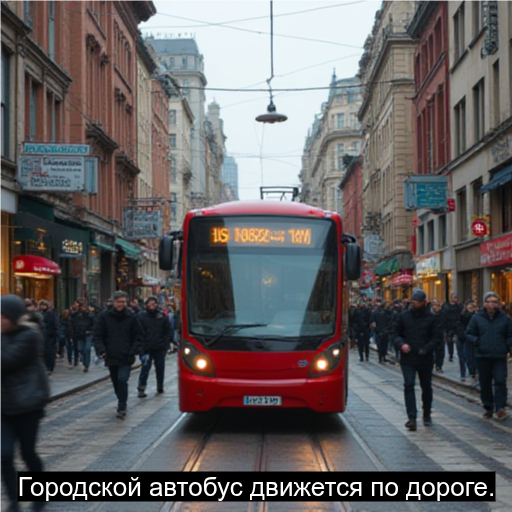}
        \caption{Ours (ru)}
    \end{subfigure}

    \caption{Images generated by FLUX text-to-image model using the prompt ``The city bus is traveling down the road'' in multiple languages (non-English captions shown on images). Our \metal-aligned model produces similar quality images compared to baseline FLUX (both T5 and CLIP encoders), FLUX-T5 and FLUX-CLIP models.}
    \label{fig:flux-base}
\end{figure*}

Qualitative analysis confirms strong semantic alignment. For the query ``A man speaks with some clicks and then loud long scrapes'', the top three retrieved audio captions were: 1) ``Sanding and filing then a man speaks'', 2) ``A man speaks with some clicking and some sanding'', and 3) ``A man speaks with a high-frequency hum with some banging and clanking''. The ground truth audio ranked 10th, but its captions—``A man talking as metal clacks followed by metal scraping against a metal surface'' and ``A man is speaking followed by saw blade noises''—closely match the top retrieved results, demonstrating robust semantic retrieval. Additional examples and details are in Appendix~\ref{app:clap}.

\section{Cross-lingual Text-to-Image Generation}
\label{sec:gen-T2I}
Our method is task-agnostic and extends naturally to generative tasks such as Text-to-Image generation. Since \metal aligns sentence-level (CLS) representations, we experiment with FLUX.1-dev (FLUX)~\cite{blackforestlabs-flux}, a 12B parameter Text-to-Image model that conditions on CLS from a CLIP encoder. FLUX is chosen for its public availability, competitive performance~\cite{yang20241}, and dual text encoders: CLIP (CLS conditioning) and T5~\cite{raffel2020exploring} (token conditioning).  

To learn the projection map $\mathcal{F}$, we align the M-MPNET encoder ($T_m$) with the CLIP encoder from FLUX ($T_e$).\footnote{In a qualitative comparison of 100 generations from FLUX$\times$LaBSE vs. FLUX$\times$M-MPNET, the latter consistently produced higher-quality images.} Since FLUX uses both CLIP and T5, we consider four variants: (i) FLUX, which inputs text to both CLIP and T5; (ii) FLUX-CLIP, which inputs text to CLIP and a generic prompt (``A photo of:'') to T5\footnote{Among tested prompts (``An image of:'', ``A picture of:'', ``A photo of:''), the last yielded best results.}; (iii) FLUX-T5, which inputs text to T5 and a generic prompt (``A photo of:'') to CLIP; and (iv) FLUX$\times$M-MPNET, which inputs text to the \metal-aligned M-MPNET encoder and a generic prompt to T5.\\

\noindent \textbf{Training Setup \& Evaluation.}  
We follow Section~\ref{sec:prelim}, training for 10 epochs without validation, using bfloat16 precision and MSE loss (instead of qq.~\ref{eq:final loss}) on unnormalized representations. Unlike retrieval tasks, $\mathcal{L}_{str}$ degrades performance: preserving scale information (exact mapping of representations) rather than structural similarity is more important for generation. Images are generated at $512{\times}512$ resolution with guidance scale 3.5, 10 inference steps, and a fixed seed. Following prior work~\cite{Ramesh2021ZeroShotTG, Rombach2021HighResolutionIS, Saharia2022PhotorealisticTD}, we sample 30K captions from MSCOCO2014~\cite{Lin2014MicrosoftCC} for English validation set. For multilingual evaluation, we extend captions into 9 languages (fr, el, he, id, ko, fa, ru, es, hi) using IndicTrans2 (hi) and Aya-23-35B (others). Metrics used are FID~\cite{Heusel2017GANsTB} and Inception Score (IS)~\cite{Salimans2016ImprovedTF}. \\

\begin{figure*}[t]
    \centering
    \begin{subfigure}{0.157\textwidth}
        \includegraphics[width=\linewidth]{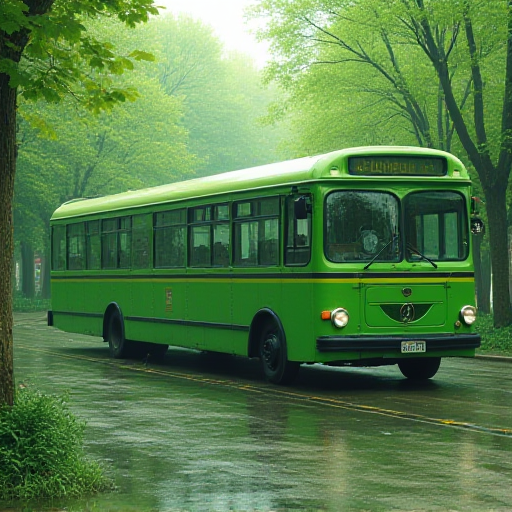}
        \caption{Ours (hi)}
    \end{subfigure}
    \hspace*{\fill}
    \begin{subfigure}{0.157\textwidth}
        \includegraphics[width=\linewidth]{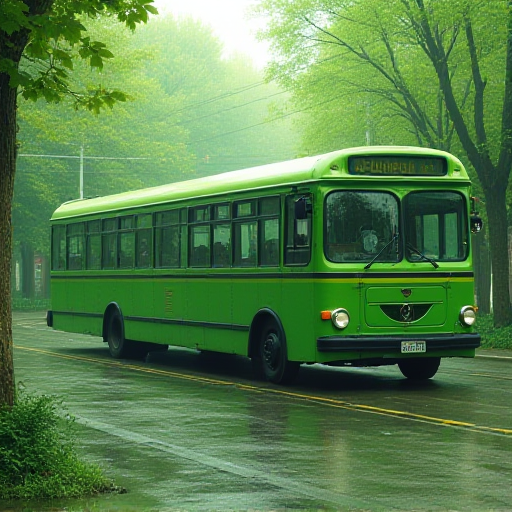}
        \caption{Ours (fr)}
    \end{subfigure}
    \hspace*{\fill}
    \begin{subfigure}{0.157\textwidth}
        \includegraphics[width=\linewidth]{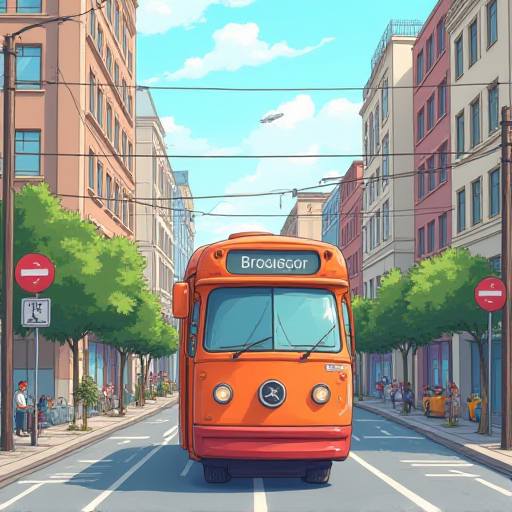}
        \caption{Ours (hi)}
    \end{subfigure}
    \hspace*{\fill}
    \begin{subfigure}{0.157\textwidth}
        \includegraphics[width=\linewidth]{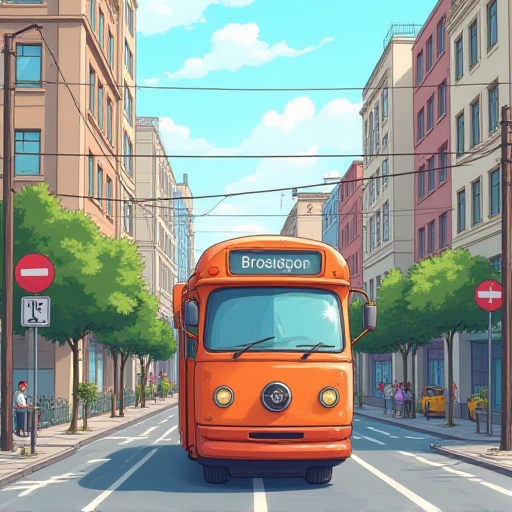}
        \caption{Ours (fr)}
    \end{subfigure}
    \hspace*{\fill}
    \begin{subfigure}{0.157\textwidth}
        \includegraphics[width=\linewidth]{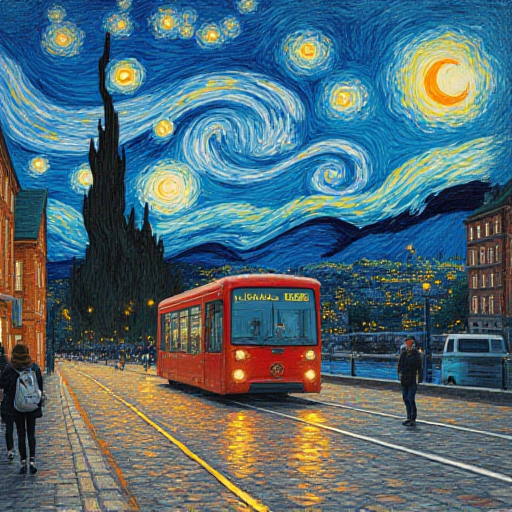}
        \caption{Ours (hi)}
    \end{subfigure}
    \hspace*{\fill}
    \begin{subfigure}{0.157\textwidth}
        \includegraphics[width=\linewidth]{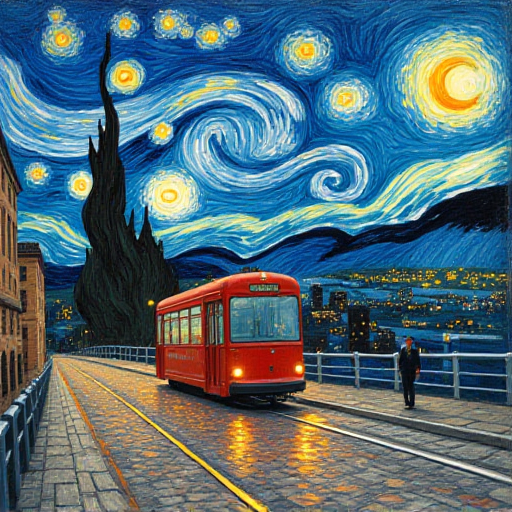}
        \caption{Ours (fr)}
    \end{subfigure}

    \vspace{10pt} 

    \begin{subfigure}{0.157\textwidth}
        \includegraphics[width=\linewidth]{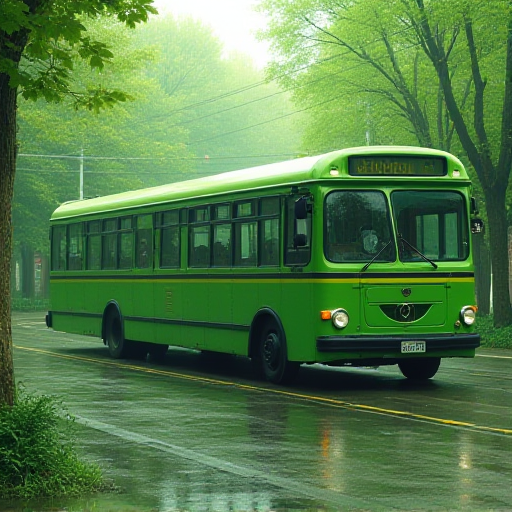}
        \caption{Ours (ru)}
    \end{subfigure}
    \hspace*{\fill}
    \begin{subfigure}{0.157\textwidth}
        \includegraphics[width=\linewidth]{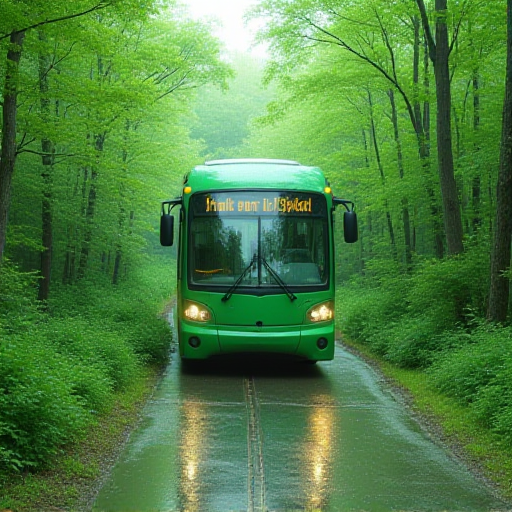}
        \caption{Ours (fa)}
    \end{subfigure}
    \hspace*{\fill}
    \begin{subfigure}{0.157\textwidth}
        \includegraphics[width=\linewidth]{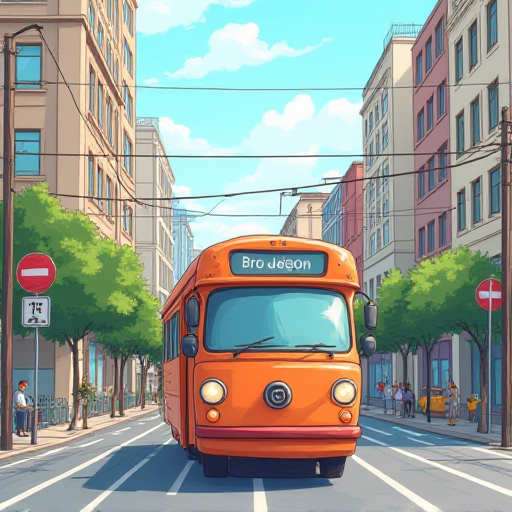}
        \caption{Ours (ru)}
    \end{subfigure}
    \hspace*{\fill}
    \begin{subfigure}{0.157\textwidth}
        \includegraphics[width=\linewidth]{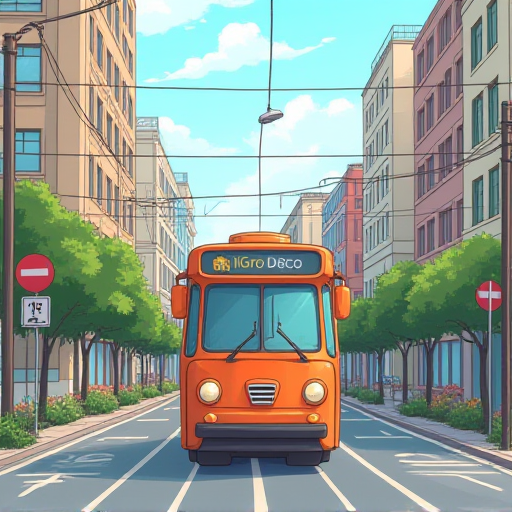}
        \caption{Ours (fa)}
    \end{subfigure}
    \hspace*{\fill}
    \begin{subfigure}{0.157\textwidth}
        \includegraphics[width=\linewidth]{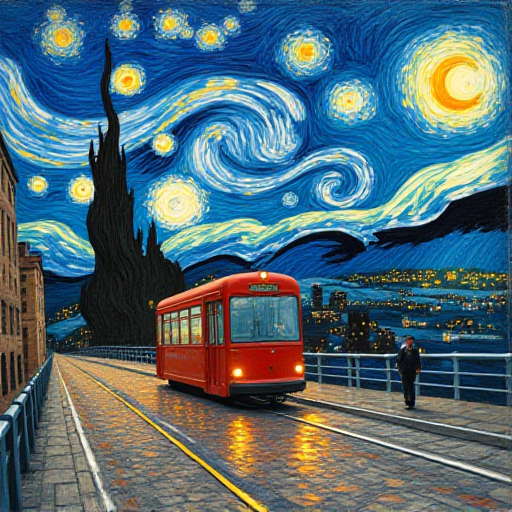}
        \caption{Ours (ru)}
    \end{subfigure}
    \hspace*{\fill}
    \begin{subfigure}{0.157\textwidth}
        \includegraphics[width=\linewidth]{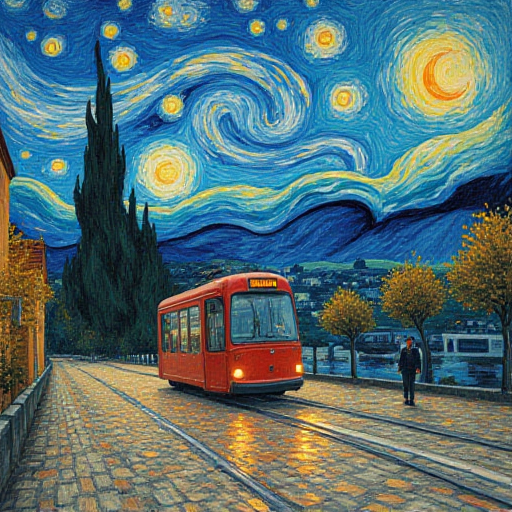}
        \caption{Ours (fa)}
    \end{subfigure}

    \hspace*{\fill}
    \begin{minipage}{0.31\textwidth}
        \centering
        {\normalfont ``A green-themed photo of: ''}
    \end{minipage}
    \hspace*{\fill}
    \begin{minipage}{0.31\textwidth}
        \centering
        {\normalfont ``A cartoon photo of: ''}
    \end{minipage}
    \hspace*{\fill}
    \begin{minipage}{0.31\textwidth}
        \centering
        {\normalfont ``A Van Gogh-style photo of: ''}
    \end{minipage}
    \hspace*{\fill}

    \caption{Images generated from multilingual translations of input prompt: ``The city bus is traveling down the road'' using FLUX $\times$ M-MPNET model, with theme prompts in T5 encoder to enhance image quality and style.}
    \label{fig:flux-theme}
\end{figure*}

\noindent \textbf{Results \& Analysis.}  
FLUX$\times$M-MPNET achieves a strong Inception Score of 31.81 averaged across languages, including 35.9$\pm$0.57 on English, surpassing trained models such as LDM~\cite{Rombach2021HighResolutionIS} (30.29$\pm$0.42), CogView~\cite{ding2021cogview} (18.2), and LAFITE~\cite{zhou2022towards} (26.02). However, FID is poor: 40.9 (FLUX-CLIP) and 43.4 (FLUX$\times$M-MPNET), compared to 23.4 for both FLUX and FLUX-T5. The identical FID for FLUX and FLUX-T5 suggests FLUX relies heavily on T5 token conditioning and can generate high-quality images without CLIP input. Since our setup replaces CLIP with an aligned encoder and uses a generic T5 prompt, generated images are less faithful to the text (e.g., missing objects).  

Despite this limitation, qualitative results (Fig.~\ref{fig:flux-base}) show diverse, semantically relevant images with slightly reduced fidelity. For FLUX-CLIP and FLUX$\times$M-MPNET, we also observe \emph{hallucinated} generations—well-formed but text-misaligned outputs. These are not random noise but coherent, object-rich images, likely caused by weak signal from T5 due to generic prompts. Adding more specific object/style cues to T5 input alleviates this, as illustrated in Fig.~\ref{fig:flux-theme}. Language-wise FID and IS score breakdowns and examples are in Appendix~\ref{app:text-to-image}.

\section{Conclusion}
We introduce \metal, an efficient method to align multilingual latent spaces with multimodal spaces using only a few linear layers and English text data. Unlike existing approaches requiring large-scale multilingual or multimodal corpora, \metal reduces resource needs while maintaining strong performance across tasks and modalities.  

On XTD-T2I retrieval, it achieves 94.9\% Recall@10 for English and 89.5\% averaged across 11 languages, demonstrating robust zero-shot transfer. Qualitative analyses, including t-SNE visualizations, show projected multilingual embeddings align closely with multimodal representations. Beyond image-text retrieval, \metal generalizes to Audio-Text retrieval and cross-lingual Text-to-Image generation. We release synthetic evaluation datasets: AudioCaps and Clotho extended to 33 languages, and MSCOCO-30K captions extended to 9 languages, providing a unified open benchmark. While promising, further improvements are possible, particularly via token-level alignment. Overall, \metal shows that lightweight, data-efficient strategies can bridge multilingual and multimodal spaces by leveraging implicit alignment between languages and modalities.

\section*{Acknowledgements}
The authors thank Prof.\ Preethi Jyothi (Indian Institute of Technology, Bombay) for her continued support and for fostering the research environment that enabled this research. Her guidance and encouragement were instrumental throughout the course of this research. We are grateful to Pavan Kalyan for extensive discussions, careful proofreading, and thoughtful feedback on early drafts of the paper, which greatly improved the clarity, structure, and presentation of this work, and to Isha Pandey for her help in reviewing, refining, and strengthening the rebuttal responses.

\pagebreak
\section*{Limitations}
\noindent \textbf{Need for local alignment.} Our method focuses on aligning \emph{global}, sentence-level representations in encoder-based models and demonstrates strong performance in this setting. However, in its current form, it does not provide alignment at the token level and therefore does not directly extend to Multimodal Large Language Models (MLLMs), where effective generation relies on fine-grained representations. Tasks such as text-to-image generation and cross-lingual skill transfer would benefit from token-level alignment signals alongside high-level semantic consistency. Extending the proposed framework to support local alignment is a natural and promising direction for future work.
\\\\
\noindent \textbf{Joint Cross-modal Representations.} Our work effectively aligns multilingual and multimodal representations from dual encoder models, where each modality is encoded individually. Joint cross-modal encoders generate representations by combining multiple modality representations through shared architectural components. The effectiveness of our method for joint cross-modal representations remains to be explored.
\\\\
\noindent \textbf{Lack of Human-verified multilingual-multimodal evaluation set.} Finding high-quality standard multilingual evaluation sets for Audio-Text retrieval and Text-to-Image Generation tasks is challenging. To address this, we curated synthetic parallel evaluation data for AudioCaps (160K samples), Clotho (172K samples), and MSCOCO-30K (270K samples). Due to the large scale of the data, human verification of the translated captions was not feasible for us. While we use objective metrics like spBLEU and chrF++ to ensure dataset quality, these measures alone are not sufficient, and without human verification, some errors may persist in the evaluation dataset.

\bibliography{custom}

\appendix
\section{Potential Risks}
There has been investigation of various biases (gender, race, etc.) for multimodal models primarily in English language. Our method extends the capability of multimodal models to many languages including low resource languages. However, there have been very few works to detect and mitigate biases for these languages. Additionally, since we use English language as anchor it is possible that the biases present in English multimodal model can manifest in the resulting multilingual multimodal model.

\section{Model \& Data License}
\label{app:license}
All models that are taken from sentence-transformers\footnote{\url{https://www.sbert.net/}} library (Multilingual CLIP (MCLIP-ST), Multilingual MPNET (M-MPNET), Multilingual MiniLM (M-MiniLM)), LaBSE, KakaoBrain-ALIGN, Jina-CLIP-v1, and LAION-CLAP (CLAP-General, CLAP-HTSAT-Fused) are under Apache License 2.0. For model FLUX.1-dev, generated outputs can be used for personal, scientific, and commercial purposes as described in the \href{https://huggingface.co/black-forest-labs/FLUX.1-dev/blob/main/LICENSE.md}{FLUX.1 [dev] Non-Commercial License}. Multilingual CLIP~\cite{carlsson-etal-2022-cross}, OpenAI-CLIP, and IndicTrans2 are under MIT License. Jina-CLIP-v2, Jina-embeddings-v3, AYA-23-35B are under CC-by-NC-4.0. Use of any combination of the models aligned using our method must adhere to the license of all individual models.

We release our extended datasets in new languages for AudioCaps, Clotho, and MSCOCO2014-30K under CC-By-NC-4.0 License, adhering to source dataset licenses and models used to generate data (AudioCaps- MIT License, Clotho- \href{https://github.com/audio-captioning/clotho-dataset?tab=License-1-ov-file#readme}{Tampere University License (non-commercial with attribution)}, MSCOCO- CC-By-4.0).

\section{List supported languages for multilingual and/or multimodal models}
\label{app:list-lang}
\begin{table*}[!htbp]
\centering
\fontsize{7pt}{9.75pt}\selectfont
\begin{tabular}{p{7.2cm} p{6.5cm}}
\toprule
Models & Supported languages \\ \midrule
LaBSE~\cite{Feng2020LanguageagnosticBS} & af, ht, pt, am, hu, ro, ar, hy, ru, as, id, rw, az, ig, si, be, is, sk, bg, it, sl, bn, ja, sm, bo, jv, sn, bs, ka, so, ca, kk, sq, ceb, km, sr, co, kn, st, cs, ko, su, cy, ku, sv, da, ky, sw, de, la, ta, el, lb, te, en, lo, tg, eo, lt, th, es, lv, tk, et, mg, tl, eu, mi, tr, fa, mk, tt, fi, ml, ug, fr, mn, uk, fy, mr, ur, ga, ms, uz, gd, mt, vi, gl, my, wo, gu, ne, xh, ha, nl, yi, haw, no, yo, he, ny, zh, hi, or, zu, hmn, pa, hr, pl \\
Jina-CLIP-v2~\cite{Koukounas2024jinaclipv2MM}, Jina-Text-v3~\cite{sturua2024jina} & ar, bn, zh, da, nl, en, fi, fr, ka, de, el, hi, id, it, ja, ko, lv, no, pl, pt, ro, ru, sk, es, sv, th, tr, uk, ur, vi \\
Multilingual CLIP~\cite{carlsson-etal-2022-cross}- LaBSE ViT-L/14, XLM-R-Large ViT-B/32, XLM-R ViT-L/14, XLM-R-Large ViT-B/16+ & af, am, ar, az, bg, bn, bs, ca, cs, cy, da, de, el, en, es, et, fa, fa-AF, fi, fr, gu, ha, he, hi, hr, ht, hu, hy, id, is, it, ja, ka, kk, kn, ko, lt, lv, mk, ml, mn, ms, mt, nl, no, pl, ps, pt, ro, ru, si, sk, sl, so, sq, sr, sv, sw, ta, te, th, tl, tr, uk, ur, uz, vi, zh, zh-TW \\
M-MPNET, M-MiniLM, MCLIP-ST~\cite{reimers-2020-multilingual-sentence-bert} & ar, bg, ca, cs, da, de, el, en, es, et, fa, fi, fr, fr-ca, gl, gu, he, hi, hr, hu, hy, id, it, ja, ka, ko, ku, lt, lv, mk, mn, mr, ms, my, nb, nl, pl, pt, pt-br, ro, ru, sk, sl, sq, sr, sv, th, tr, uk, ur, vi, zh-cn, zh-tw, zh \\
\bottomrule
\end{tabular}
\caption{List of Multilingual text encoder and multilingual multimodal models and it's supported languages.}
\label{tab:supp-langs}
\end{table*}
Different multilingual text encoder and multilingual CLIP models support different languages. For fairer comparison, we also report metrics averaged on model-supported languages (e.g. Table~\ref{tab:clap} and Table~\ref{tab:main-supported-lang}). Table~\ref{tab:supp-langs} shows a list of models and their supported languages.

\section{Ablations experiments on Image-text Retrieval}
\label{app:prelim}
Table~\ref{tab:arch:T2I} and ~\ref{tab:arch:I2T} shows our method outperforms all other training objectives on Text-to-Image retrieval for XTD dataset. The impact of high $\lambda$ is significant for both Text-to-Image retrieval and Image-to-Text retrieval (0.5\% gain in Avg. Recall@10) shown in Table~\ref{tab:arch:I2T} and Table~\ref{tab:arch:T2I}.
\begin{table*}[!htbp]
\centering
\fontsize{7pt}{9.75pt}\selectfont
\begin{tabular}{p{2.5cm} c c p{0.25cm} p{0.25cm} p{0.25cm} p{0.25cm} p{0.25cm} p{0.25cm} p{0.25cm} p{0.25cm} p{0.25cm} p{0.25cm} p{0.25cm} p{0.25cm}}
\toprule

Loss & MLP layers & Skip Conn. & Avg. & de & en & es & fr & it & jp & ko & pl & ru & tr & zh \\ \midrule
MSE & 2 & No & 89 & 90.2 & 94.3 & 90.8 & 90.1 & 90.4 & 82 & 85.5 & 90.8 & 86 & 88.9 & 89.5 \\
MSE & 2 & Yes & 88.9 & 89.9 & 94.2 & 90.3 & 90 & 90.7 & 82 & 85.4 & 91 & 85.9 & 88.6 & 89.4 \\
$\lambda_1$*$\mathcal{L}_{align}$ + $\beta_1$*$\mathcal{L}_{str}$ & 2 & No & \textbf{89.5} & 90.5 & 94.7 & \textbf{91.9} & 90.5 & 91.1 & 82.4 & \textbf{85.8} & 91.2 & \textbf{86.8} & \textbf{89.2} & 89.9 \\
$\lambda_1$*$\mathcal{L}_{align}$ + $\beta_1$*$\mathcal{L}_{str}$ & 2 & Yes & 89.2 & 90.1 & 94.3 & 90.6 & 90.6 & 91.1 & 82.5 & 85.4 & 91.1 & 86.6 & 88.2 & \textbf{90.3} \\
$\lambda_1$*$\mathcal{L}_{align}$ + $\beta_1$*$\mathcal{L}_{str}$ & 4 & No & 89.1 & 90.3 & 94.3 & 90.8 & \textbf{90.7} & 90.8 & 82.1 & 85.5 & 90.7 & 86.4 & 88.4 & 90.2 \\
$\lambda_1$*$\mathcal{L}_{align}$ + $\beta_1$*$\mathcal{L}_{str}$ & 1 & No & 89.3 & \textbf{90.6} & 94.6 & 90.8 & 90 & 90.9 & \textbf{82.7} & 85.7 & 91 & 86.6 & 89 & 90.1 \\
Similarity Loss & 2 & No & 88.8 & 89.7 & 94.2 & 90.4 & 89.6 & 90.5 & 81.6 & 84.9 & 90.8 & 86 & 89.2 & 89.7 \\
L1 & 2 & No & 86.7 & 87.7 & 94.2 & 89.2 & 89.4 & 89.1 & 76.9 & 81.1 & 88.7 & 83.7 & 86.1 & 87.6 \\
$\lambda_2$*$\mathcal{L}_{align}$ + $\beta_1$*$\mathcal{L}_{str}$ & 2 & No & 89 & 89.9 & \textbf{94.8} & 90.8 & 89.8 & \textbf{91.4} & 82.3 & 84.7 & \textbf{91.5} & 85.7 & 88.2 & 89.7 \\

\bottomrule
\end{tabular}
\caption{Comparison of Recall@10 metric across different training losses, and settings- varying number of linear layers, presence or absence of residual connections (Skip Conn.) between linear layers for \metal-aligned Jina-CLIP-v1$\times$M-MPNET on XTD dataset for Text-to-Image retrieval task. $\lambda_1=48, \lambda_2=1, \beta_1=1$.}
\label{tab:arch:T2I}

\end{table*}

\begin{table*}[!htbp]
\centering
\fontsize{7pt}{9.75pt}\selectfont
\begin{tabular}{p{2.5cm} c c p{0.25cm} p{0.25cm} p{0.25cm} p{0.25cm} p{0.25cm} p{0.25cm} p{0.25cm} p{0.25cm} p{0.25cm} p{0.25cm} p{0.25cm} p{0.25cm}}
\toprule

Loss & MLP layers & Skip Conn. & Avg. & de & en & es & fr & it & jp & ko & pl & ru & tr & zh \\ \midrule
MSE & 2 & No & 88.7 & 89.2 & 95.3 & 89.6 & 89.6 & 90 & 81.7 & 85.4 & 90.2 & 85.5 & 89.4 & 89.9 \\
MSE & 2 & Yes & 88.6 & 89.4 & 95.5 & 89.8 & 89.2 & 89.4 & 81.9 & 84.9 & 90.1 & 85.5 & 89.2 & 89.7 \\
$\lambda_1$*$\mathcal{L}_{align}$ + $\beta_1$*$\mathcal{L}_{str}$ & 2 & No & \textbf{89.4} & \textbf{89.6} & 95.2 & 91.2 & \textbf{89.7} & 90.4 & \textbf{83.1} & \textbf{85.9} & 90.8 & 87 & 90.1 & 90.3 \\
$\lambda_1$*$\mathcal{L}_{align}$ + $\beta_1$*$\mathcal{L}_{str}$ & 2 & Yes & \textbf{89.4} & 89.2 & 95.5 & 90.8 & 89.5 & 90.5 & 83 & 85.5 & 91 & \textbf{87.2} & \textbf{90.3} & \textbf{90.4} \\
$\lambda_1$*$\mathcal{L}_{align}$ + $\beta_1$*$\mathcal{L}_{str}$ & 4 & No & 89.2 & 89.3 & 95.2 & 90.7 & 89.5 & 90.5 & 83 & 85.1 & 90.9 & 86.8 & 89.8 & \textbf{90.4} \\
$\lambda_1$*$\mathcal{L}_{align}$ + $\beta_1$*$\mathcal{L}_{str}$ & 1 & No & 89.3 & 89.5 & \textbf{95.8} & 91 & 89.4 & \textbf{90.6} & 82.4 & 85.3 & \textbf{91.1} & 86.5 & 90.1 & 90.2 \\
Similarity Loss & 2 & No & 88.7 & 89.2 & 95.7 & 90 & 89.1 & 89 & 81.5 & 85.4 & 90.5 & 85.5 & 89.3 & 90.1 \\
L1 & 2 & No & 83.9 & 85.6 & 95 & 86.2 & 84.9 & 87.1 & 74.9 & 78.5 & 84.1 & 78.7 & 83.2 & 85.1 \\
$\lambda_2$*$\mathcal{L}_{align}$ + $\beta_1$*$\mathcal{L}_{str}$ & 2 & No & 88.9 & 89.2 & 95.3 & \textbf{91.3} & \textbf{89.7} & \textbf{90.6} & 82.4 & 84.3 & 90.2 & 86 & 89.4 & 90 \\

\bottomrule
\end{tabular}
\caption{Comparison of Recall@10 metric across different training losses, and settings- varying number of linear layers, presence or absence of residual connections (Skip Conn.) between linear layers for \metal-aligned Jina-CLIP-v1$\times$M-MPNET on XTD dataset for Image-to-Text retrieval task. $\lambda_1=48, \lambda_2=1, \beta_1=1$.}
\label{tab:arch:I2T}

\end{table*}

\section{Weight Analysis}
\label{app:weight analysis}

To further investigate the behavior of the mapping, we performed a pairwise cosine distance analysis on clusters of semantically identical sentences that differ only in style or phrasing. Distances were computed for (i) Jina-CLIP-v1 embeddings, (ii) multilingual MPNET embeddings, and (iii) our mapped embeddings. All embeddings were $\ell_2$-normalized, and only the upper-triangle of the distance matrix was considered to avoid double-counting symmetric pairs. Results show that Jina-CLIP-v1 embeddings exhibit the largest variability (0.028--0.083), reflecting sensitivity to small syntactic or stylistic changes.fig:appendix boxplots Multilingual MPNET embeddings are more compact (0.011--0.043), consistent with their language-agnostic nature, while our mapped embeddings fall in between (0.011--0.046), indicating successful alignment to the Jina-CLIP-v1 space while preserving semantic consistency. These observations support our hypothesis that Jina-CLIP-v1 text encoders capture language-specific features absent in language-agnostic embeddings like M-MPNET, and that the learned linear map effectively projects onto the relevant subspace for retrieval.

\begin{figure}[!htbp]
    \centering
    \includegraphics[width=\columnwidth]{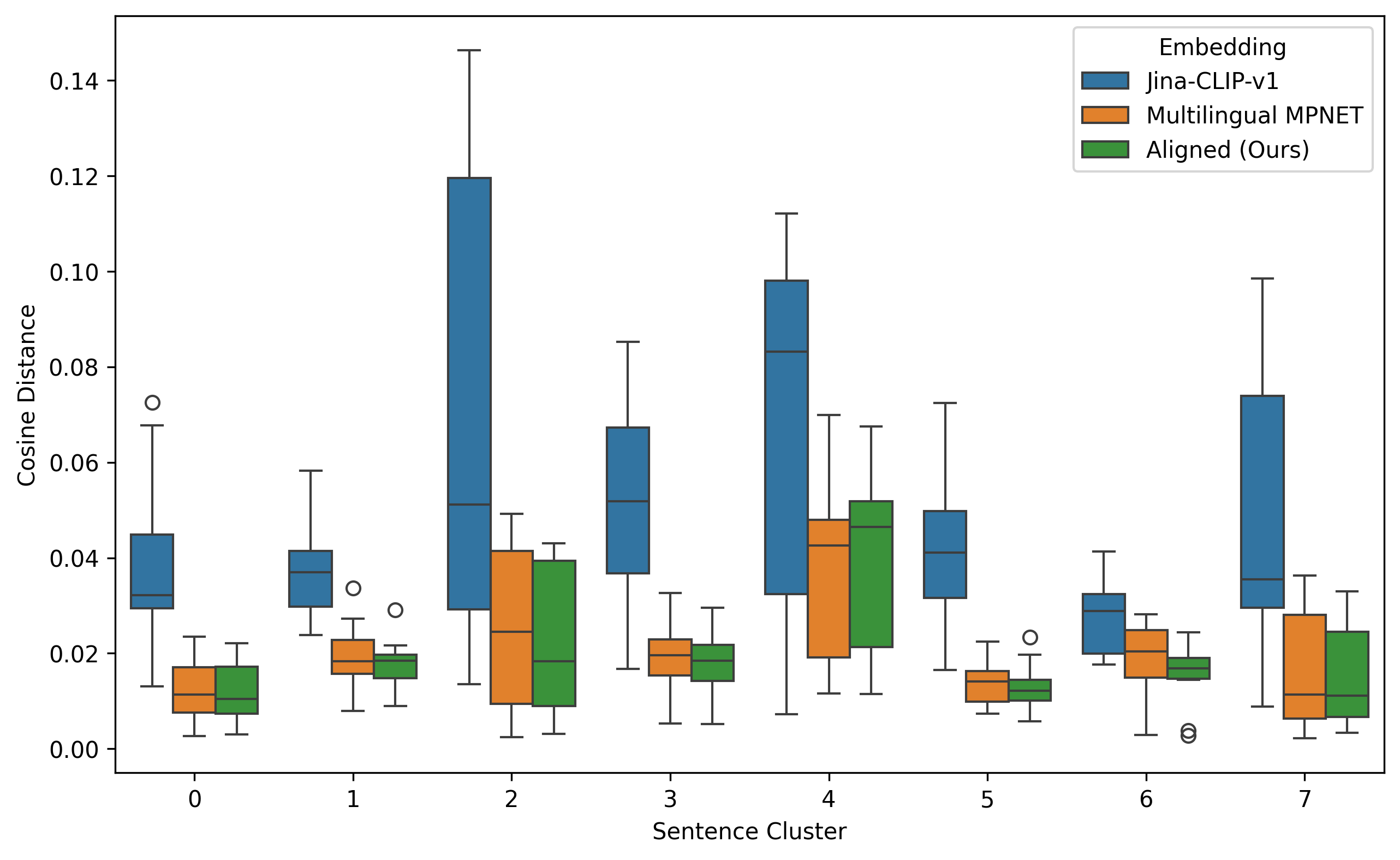}
    \caption{Pairwise cosine distances for clusters of semantically identical sentences with stylistic or syntactic variations. Jina-CLIP-v1 embeddings (blue) show the largest variability, reflecting sensitivity to language-specific phrasing. Multilingual MPNET embeddings (orange) are more compact, consistent with language-agnostic representations, and our mapped embeddings (green) fall in between, indicating successful alignment while preserving semantic consistency. Each boxplot summarizes the distribution of distances within a cluster.}
    \label{fig:appendix boxplots}
\end{figure}

Figure~\ref{fig:appendix boxplots} presents boxplots for all clusters, illustrating the distribution of pairwise distances and confirming the trends described above. Below, we list the clusters of nearly identical sentences used for this analysis.

\subsection{Sentence Variation Clusters}
\begin{itemize}
    \item \textbf{Dog / Animal}
    \begin{itemize}
        \item A dog running in the park.
        \item The dog is running in the park.
        \item A dog runs in the park.
        \item A dog is running through the park.
        \item In the park, a dog runs.
    \end{itemize}

    \item \textbf{Cat}
    \begin{itemize}
        \item A cat sleeps on the sofa.
        \item The cat is sleeping on the sofa.
        \item A sleeping cat is on the sofa.
        \item On the sofa, a cat is sleeping.
        \item A cat is lying asleep on the sofa.
    \end{itemize}

    \item \textbf{Human Actions: Cycling}
    \begin{itemize}
        \item A person is riding a bicycle on the street.
        \item A person is riding a bike along the street.
        \item A cyclist is riding on the street.
        \item The person rides a bicycle on the street.
        \item On the street, a person is riding a bicycle.
    \end{itemize}

    \item \textbf{Human Actions: Drawing}
    \begin{itemize}
        \item A child is drawing on a piece of paper.
        \item The child is drawing on a sheet of paper.
        \item On paper, a child is drawing.
        \item A child is drawing on paper.
        \item The child draws on a piece of paper.
    \end{itemize}

    \item \textbf{Nature / Scenery: Sunset}
    \begin{itemize}
        \item A sunset over the mountains.
        \item The sun is setting over the mountains.
        \item The sun is setting near the mountains.
        \item The mountains during sunset.
        \item The setting sun is over the mountains.
    \end{itemize}

    \item \textbf{Nature / Scenery: River}
    \begin{itemize}
        \item A river flows through the forest.
        \item The river is flowing through the forest.
        \item Through the forest flows a river.
        \item The forest has a river flowing through it.
        \item In the forest, a river flows.
    \end{itemize}

    \item \textbf{Objects / Still Life: Car}
    \begin{itemize}
        \item A red car is parked on the street.
        \item On the street, a red car is parked.
        \item A car is parked on the street, and the car is red.
        \item A car colored red is parked on the street.
        \item A red-colored car is parked on the street.
    \end{itemize}

    \item \textbf{Objects / Still Life: Coffee}
    \begin{itemize}
        \item A cup of coffee is on the table.
        \item On the table is a cup of coffee.
        \item The cup of coffee is on the table.
        \item There is a coffee cup on the table.
        \item A mug of coffee sits on the table.
    \end{itemize}
\end{itemize}

\section{Image-Text Retrieval: Additional Results}
\label{app:image-text}
\subsection{Language-wise Recall on XM3600 \& Multi30K}
\begin{table*}[!htbp]
    \centering
    \fontsize{7pt}{9.75pt}\selectfont
    \begin{tabular}{c c c c c c c c c c c c c}
    \toprule
        Retrieval Type & Avg & ar & bn & cs & da & de & el & en & es & fa & fi & fil \\ \midrule
        T2I & 66.3 & 68.3 & 31.3 & 73 & 79.3 & 80.9 & 67.3 & 79.8 & 75.7 & 76 & 77.3 & 9.8 \\
        I2T & 73.1 & 77.4 & 36.6 & 80.2 & 86.6 & 88 & 76.1 & 84.9 & 82.4 & 82 & 84.8 & 17.9 \\

        \bottomrule
    \end{tabular}

    \vspace{10pt} 

    
    \begin{tabular}{c c c c c c c c c c c c c}
    \toprule
Retrieval Type & fr & he & hi & hr & hu & id & it & ja & ko & mi & nl & no \\ \midrule
    T2I & 81.4 & 74.7 & 60 & 79.4 & 77.7 & 85.5 & 79.4 & 78.7 & 72.4 & 0.7 & 75 & 78.9 \\
    I2T & 88.3 & 82.6 & 70.7 & 87.3 & 83.5 & 90 & 85.8 & 85.6 & 81.2 & 1.1 & 80 & 86.9 \\
    \bottomrule
    \end{tabular}

    \vspace{10pt}

    \begin{tabular}{c c c c c c c c c c c c c c}
    \toprule
Retrieval Type & pl & pt & quz & ro & ru & sv & sw & te & th & tr & uk & vi & zh \\ \midrule
    T2I & 76.2 & 77.2 & 2.7 & 80 & 82.2 & 78 & 4.5 & 28.8 & 78.9 & 74.9 & 76.8 & 81.9 & 80.9 \\
    I2T & 83.6 & 84.1 & 6.1 & 87.6 & 88.6 & 85.3 & 9.1 & 39.1 & 86.1 & 81.4 & 84 & 89.2 & 87.4 \\
    \bottomrule
    \end{tabular}
    
    \caption{Recall@10 across 36 languages for XM3600 on I2T and T2I retrieval task using \metal-aligned Jina-CLIP-v1 $\times$ M-MPNET.}
    \label{tab:xm3600}
\end{table*}
\begin{table*}[!htbp]
\centering
\fontsize{7pt}{9.75pt}\selectfont
\begin{tabular}{l c c c c c c c c c c}
\toprule
\multirow{2}{*}{Model} & \multicolumn{5}{c}{T2I} & \multicolumn{5}{c}{I2T} \\
 & Avg & cs & de & en & fr & Avg & cs & de & en & fr \\ \midrule

Jina-CLIP-v1 $\times$ M-MPNET & 89.9 & 87.9 & \textbf{89.5} & 91.8 & 90.3 & 89.7 & 86.8 & 89.7 & 91.7 & 90.4 \\
CLIP $\times$ M-MPNET & \textbf{90.1} & \textbf{88.1} & 88.3 & \textbf{93.2} & \textbf{90.8} & \textbf{92.1} & \textbf{90} & \textbf{90.9} & \textbf{95.1} & \textbf{92.3} \\

\bottomrule
\end{tabular}
    \caption{Recall@10 across 4 languages for Multi30K on I2T and T2I retrieval task using \metal-aligned Jina-CLIP-v1 $\times$ M-MPNET.}
\label{tab:multi30k}

\end{table*}
Tables~\ref{tab:xm3600}, \ref{tab:multi30k} show language-wise performance of our \metal-aligned models on XM3600 and Multi30K datasets respectively. Interestingly, CLIP $\times$ M-MPNET outperforms Jina-CLIP-v1 $\times$ M-MPNET by 2.4\% I2T and 0.2\% T2I on Multi30K dataset. 

\subsection{Results on model-supported languages}
\begin{table*}[!htbp]
\centering
\fontsize{7pt}{9.75pt}\selectfont
\begin{tabular}{l c c c c}
\toprule
\multirow{2}{*}{Models} & \multicolumn{2}{c}{XM3600} & \multicolumn{2}{c}{Multi30K} \\ 
 & T2I & I2T & T2I & I2T \\ \midrule
\multicolumn{5}{l}{\textbf{English-only Zero-shot Baseline Models}\vspace{0.3em}}\\ 
E1: CLIP (ViT-L 336px) & 77.3 & 87.1 & 93.5 & \textbf{95.8} \\
E2: Jina-CLIP-v1 & 85.7 & 91.8 & 93.5 & 93.6 \\
E3: K-ALIGN & \textbf{87.0} & \textbf{92.0} & \textbf{95.9} & \textbf{95.8} \\
\midrule
\multicolumn{5}{l}{\textbf{Multilingual Multimodal Models Trained on}}  \\
\multicolumn{5}{l}{\textbf{Supervised Multimodal and/or Multilingual Data}\vspace{0.3em}}  \\

T2: MCLIP-ST & 57.6 & 71.1 & 80.7 & 83.4 \\
T5: LABSE ViT-L/14 & 77.0 & 87.5 & 90.9 & 93.7 \\
T6: XLM-R-L ViT-B/32 & 79.6 & 89.0 & 89.2 & 91.0 \\
T7: XLM-R ViT-L/14 & 80.9 & 89.6 & 92.2 & 94.4 \\
T8: XLM-R-L ViT-B/16+ & 86.5 & 91.9 & 93.9 & 94.2 \\
T9: Jina-CLIP-v2 & \textbf{90.1} & \textbf{93.9} & \textbf{94.3} & \textbf{94.5} \\ \midrule

\multicolumn{5}{l}{\textbf{\metal-aligned Multilingual Multimodal models}} \\
\multicolumn{5}{l}{\textbf{Trained on only English Text data}\vspace{0.3em}} \\

M1: Jina-CLIP-v1 $\times$ LaBSE & 64.9 &	67.5	& 79	& 75.7 \\
M2: Jina-CLIP-v1 $\times$ M-MiniLM & 68.5	& 75.7	& 88	& 85.9 \\
M3: Jina-CLIP-v1 $\times$ JinaTextV3 & 75.4	& 80.1	& 87.8	& 87.5 \\
M4: Jina-CLIP-v1 $\times$ M-MPNET & \textbf{76.8}	& \textbf{84.0} &	89.9	& 89.7 \\
M5: CLIP $\times$ M-MPNET & 65.4	& 77.6	& 90.1	& \textbf{92.1} \\
M6: K-ALIGN  $\times$ M-MPNET & 68.6	& 78.3	& \textbf{91.0}	& 90.2 \\
\bottomrule
\end{tabular}
\caption{Performance of \metal-align models in comparison with English and Mutlingual CLIP-like models on Recall@10 metric for supported languages for XM3600 and Multi30K datasets.}
\label{tab:main-supported-lang}
\end{table*}
Similar to our results for Image-Text retrieval in Table~\ref{tab:main}, in Table~\ref{tab:main-supported-lang}, we report Recall@10 metric averaged only on languages supported by the respective multilingual text encoder/mutlilingual CLIPs. Supported languages for each model is listed in Table~\ref{tab:supp-langs}. 

\subsection{Reproducibility experiments}
To show that our method's performance is reproducible. We run experiments twice on our method for Image-Text retrieval task, and report mean and standard deviation in Tables~\ref{tab:std:IT} to show that the performance is stable across varying random seeds. 

\begin{table*}[!htbp]
\centering
\fontsize{7pt}{9.75pt}\selectfont
\setlength{\tabcolsep}{0.5em}
\begin{tabular}{l c c c c c c c c c c c c}
\toprule
Retr. Type & Avg. & de & en & es & fr & it & jp & ko & pl & ru & tr & zh \\ \midrule
T2I & 89.4$\pm$0.1 & 90.6$\pm$0.1 & 94.6$\pm$0.1 & 91.5$\pm$0.6 & 90.4$\pm$0.1 & 91.1$\pm$0.0 & 82.7$\pm$0.4 & 85.5$\pm$0.4 & 91.1$\pm$0.1 & 86.8$\pm$0.1 & 89.4$\pm$0.3 & 89.8$\pm$0.1 \\
I2T & 89.4$\pm$0.1 & 89.2$\pm$0.6 & 95.4$\pm$0.2 & 91.0$\pm$0.3 & 89.6$\pm$0.1 & 90.4$\pm$0.0 & 83.1$\pm$0.0 & 85.7$\pm$0.4 & 91.0$\pm$0.3 & 87.0$\pm$0.0 & 90.1$\pm$0.0 & 90.4$\pm$0.1 \\

\bottomrule
\end{tabular}
\caption{Performance of \metal-aligned Jina-CLIP-v1 $\times$ M-MPNET on Recall@10 metrics averaged ($\pm$ standard deviation) over 2 different runs across 11 languages for Image-Text retrieval task on XTD dataset.}
\label{tab:std:IT}
\end{table*}

\section{Curation of Synthetic evaluation dataset}
\label{app:trans}
For AYA-23-35B, we use translation prompts to generate synthetic data following~\cite{alam2024maya}. We experiment with zero-shot and 3-shot prompts. We use FLoRes-200 dataset to assess the quality of translation prompts. Zero-shot prompt is fairly straightforward method- we pass the input sentence and prompt the model to generate translation in target language. For 3-shot prompt, for each input english text for which translation has to be generated, we pick 3 examples. These 3 examples are picked from sampling set- created by combining FLoRes-200 validation and test set (excluding current input text). We compute cosine similarity between input text and sampling set using LaBSE, and select top 3 texts and it's corresponding translation of the target language as a few-shot example.  The zero-shot translation prompt performs better on the FLoReS-200 dataset~\cite{costa2022no} across 14 languages\footnote{ar, zho-Hant, fr, de, he, hi, it, jp, ko, pl, ru, es, tr, vi}, achieving a mean spBLEU of 39.7 and mean chrF++ of 51.5, compared to the 3-shot prompt with mean spBLEU of 37.2 and mean chrF++ of 47.4. Given these results, we apply the zero-shot prompt to generate Aya-23-35B translations for all 22 languages. Language-wise spBLEU and chrF++ scores for AYA-23-35B are shown in Table~\ref{tab:aya-flores}, and for backtranslated Indic translations are shown in Table~\ref{tab:indictrans-eval}. Zero-shot prompt and 3-shot prompt templates are listed in Table~\ref{tab:0-shot} and \ref{tab:3-shot}.
\begin{table}[!htbp]
    \centering
    \begin{tcolorbox}[colback=blue!5!white, colframe=blue!75!black]
    You are an expert in translations. Your task is to accurately translate the following text into [target language].\\
    Input text: [input test sentence]\\
    Translation:
    \end{tcolorbox}
    
    \caption{Zero-shot prompt used generating translation from AYA-23-35B. Text in square bracket is a placeholder for actual input}
    \label{tab:0-shot}
\end{table}
\begin{table}[!htbp]
    \centering
    \begin{tcolorbox}[colback=blue!5!white, colframe=blue!75!black]
You are an expert in translations. Your task is to accurately translate the following text into [target language].
\\\\
Here are a few examples to help you understand the format:
\\\\
Example 1:\\
Input text: [input text 1]\\
Translation: [translation 1]
\\\\
Example 2:\\
Input text: [input text 2]\\
Translation: [translation 2]
\\\\
Example 3:\\
Input text: [input text 3]\\
Translation: [translation 3]
\\\\
Now, translate the following text:
\\\\
Input text: [input test sentence]\\
Translation:
\end{tcolorbox}

    \caption{3-shot prompt template used to compare effect of few-shots on translation quality for AYA-23-35B. Text in square bracket is a placeholder for actual input.}
    \label{tab:3-shot}
\end{table}
\begin{table*}[!htbp]
\centering
\fontsize{7pt}{9.75pt}\selectfont

\begin{tabular}{l c c c c c c c c c c c c c c c}
\toprule
Prompts & Avg. & ar & zh & fr & de & he & hi & it & jp & ko & pl & ru & es & tr & vi \\ \midrule
\multicolumn{16}{l}{\textbf{spBLEU}\vspace{0.3em}} \\ 
3-shot prompt & 37.2 & \textbf{46.7} & 20.6 & 67.3 & 50.0 & 37.2 & 8.9 & \textbf{64.8} & 25.4 & 16.7 & 22.6 & \textbf{56.5} & 47.5 & \textbf{36.0} & 19.9 \\
zero-shot prompt & \textbf{39.7} & 21.6 & \textbf{21.3} & \textbf{69.2} & \textbf{56.2} & \textbf{55.6} & \textbf{28.2} & 54.2 & \textbf{28.6} & \textbf{17.0} & \textbf{33.7} & 51.6 & \textbf{51.9} & 29.1 & \textbf{37.4} \\ \midrule
\multicolumn{16}{l}{\textbf{chrF++}\vspace{0.3em}} \\
3-shot prompt & 47.4 & \textbf{36.5} & \textbf{31.4} & 63.0 & 55.8 & 71.8 & 18.6 & \textbf{80.3} & 29.4 & 17.7 & 39.0 & \textbf{61.1} & 55.2 & 60.0 & 44.0 \\
zero-shot prompt & \textbf{51.5} & 29.0 & 26.3 & \textbf{64.5} & \textbf{58.6} & \textbf{77.6} & \textbf{49.0} & 78.0 & \textbf{31.4} & \textbf{22.3} & \textbf{41.0} & 58.5 & \textbf{57.8} & \textbf{63.8} & \textbf{62.7} \\
\bottomrule
\end{tabular}
\caption{spBLEU and chrF++ scores for zero-shot and 3-shot prompts for FLoRes-200 using AYA-23-35B model. zh in the table denotes Chinese Traditional (zh-Hant).}
\label{tab:aya-flores}

\end{table*}
\begin{table*}[!htbp]
\centering
\fontsize{7pt}{9.75pt}\selectfont
\begin{tabular}{l c c c c c c c c c c c c }
\toprule
Test-Dataset & Avg. & bn & gu & hi & kn & ml & mr & ne & pa & ta & te & ur   \\ \midrule
\multicolumn{13}{l}{\textbf{spBLEU}\vspace{0.3em}} \\
AudioCaps & 48.7 & 38.3 & 24.3 & 39.3 & 56.2 & 79.5 & 19.1 & 33.0 & 53.1 & 64.3 & 100.0 & 28.1 \\
CLOTHO & 47.4 & 46.3 & 51.4 & 51.4 & 31.5 & 28.7 & 67.9 & 51.4 & 48.1 & 51.4 & 43.3 & 50.4 \\ \midrule
\multicolumn{13}{l}{\textbf{chrF++}\vspace{0.3em}} \\
AudioCaps & 63.6 & 60.6 & 54.0 & 58.6 & 73.0 & 74.2 & 37.7 & 46.9 & 82.1 & 61.5 & 100.0 & 51.5 \\
CLOTHO & 59.6 & 43.8 & 64.0 & 65.9 & 46.0 & 52.4 & 68.0 & 60.5 & 65.3 & 65.9 & 58.2 & 65.2 \\
\bottomrule
\end{tabular}

\caption{spBLEU and chrF++ scores on English backtranslations of AudioCaps and Clotho dataset using IndicTrans2 models.}
\label{tab:indictrans-eval}
\end{table*}

\section{CLAP}
\label{app:clap}
\subsection{Language-wise Recall on Synthetic Evaluation Dataset.}
We show language-wise performance of \metal-aligned CLAP-general $\times$ M-MPNET on AudioCaps in Table~\ref{tab:clap-audiocaps} and Clotho in Table~\ref{tab:clap-clotho}.
\begin{table*}[!htbp]
\centering
\fontsize{7pt}{9.75pt}\selectfont

\begin{tabular}{l c c c c c c c c c c c c c}
\toprule
Retrieval Type & Avg & ar & bn & cs & de & el & en & fr & gu & he & hi & id & it \\ \midrule
T2A & 48.5 & 46.2 & 23.3 & 57.6 & 58.5 & 55.6 & 77.4 & 60.8 & 38.4 & 48.7 & 52.9 & 59 & 58.2 \\
A2T & 61 & 61.2 & 35.3 & 66.2 & 68.7 & 69.5 & 81.6 & 70.2 & 53 & 61.3 & 63.4 & 68.7 & 67.4 \\

\bottomrule
\end{tabular}
\vspace{10pt}

\begin{tabular}{l c c c c c c c c c c c c c}
\toprule
Retrieval Type & ja & kn & ko & ml & mr & nl & ne & pa & fa & pl & pt & ro & ru \\ \midrule
T2A & 50.2 & 24 & 48.5 & 26.4 & 46.3 & 60.8 & 33.8 & 22.7 & 54.6 & 52.5 & 61.3 & 58.2 & 49.8 \\
A2T & 66.4 & 39.1 & 63.3 & 42.6 & 61.9 & 70.6 & 47.1 & 35.4 & 69.7 & 66.4 & 69.8 & 68 & 64.8 \\

\bottomrule
\end{tabular}
\vspace{10pt}

\begin{tabular}{l c c c c c c c c c }
\toprule

Retrieval Type & es & ta & te & tr & uk & ur & vi & zh (hans) & zh (hant) \\ \midrule
T2A & 59.6 & 29.3 & 25.7 & 56.7 & 46.8 & 44.7 & 56.6 & 53.3 & 51.1 \\
A2T & 67.9 & 42.9 & 38.5 & 64.8 & 62.8 & 58 & 68.3 & 70.6 & 67.9 \\
\bottomrule
\end{tabular}

\caption{Recall@10 metric across 34 languages on AudioCaps dataset for Audio-to-Text (A2T) and Text-to-Audio (T2A) retrieval task using \metal-aligned CLAP-general $\times$ M-MPNET model.}
\label{tab:clap-audiocaps}
\end{table*}
\begin{table*}[!htbp]
\centering
\fontsize{7pt}{9.75pt}\selectfont
\begin{tabular}{l c c c c c c c c c c c c c}
\toprule
Retrieval Type & Avg & ar & bn & cs & de & el & en & fr & gu & he & hi & id & it \\ \midrule
T2A & 33.3 & 34.2 & 19 & 37.5 & 37.8 & 36 & 47.6 & 40.1 & 25.9 & 33.7 & 36.6 & 39.6 & 38.1 \\
A2T & 39.5 & 42 & 24.4 & 41.8 & 42.5 & 41.3 & 50.7 & 44.8 & 34.2 & 39 & 42.9 & 44.7 & 44.7 \\
\bottomrule
\end{tabular}
\vspace{10pt}

\begin{tabular}{l c c c c c c c c c c c c c}
\toprule
Retrieval Type & ja & kn & ko & ml & mr & nl & ne & pa & fa & pl & pt & ro & ru \\ \midrule
T2A & 38.8 & 17.2 & 35 & 20.9 & 29.2 & 38.6 & 23.8 & 18 & 37 & 36.6 & 39.2 & 38.3 & 35.5 \\
A2T & 45.6 & 25.5 & 43 & 27.7 & 36.5 & 45.1 & 31.3 & 25.1 & 43.3 & 43.2 & 43.9 & 43.6 & 43.8 \\
\bottomrule
\end{tabular}
\vspace{10pt}

\begin{tabular}{l c c c c c c c c c}
\toprule
Retrieval Type & es & ta & te & tr & uk & ur & vi & zh (hans) & zh (hant) \\ \midrule
T2A & 39.8 & 20.8 & 17.5 & 36.3 & 33.4 & 31.2 & 39 & 39.9 & 39.6 \\
A2T & 45.5 & 27.9 & 24.5 & 43.2 & 41.1 & 38.2 & 44.1 & 45.4 & 43.9 \\
\bottomrule
\end{tabular}

\caption{Recall@10 metric across 34 languages on Clotho dataset for Audio-to-Text (A2T) and Text-to-Audio (T2A) retrieval task using \metal-aligned CLAP-general $\times$ M-MPNET model.}
\label{tab:clap-clotho}
\end{table*}

\subsection{Quantifying the qualitative analysis and more examples}
We see in Table~\ref{tab:clap} that \metal-aligned models don't match the performance of baseline CLAP models. For English, qualitative analysis revealed that the retrieved audio for a query text had high semantic similarity. To verify our qualitative analysis, we perform following quantitative test. For each query text, we retrieve top five audios using \metal-aligned model CLAP-General $\times$ M-MPNET. Next, we compute cosine similarity between query text and captions of retrieved audio using CLAP-general model. On average, we see higher cosine similarity for CLAP-general $\times$ M-MPNET (0.7) compared to CLAP-general (0.65), demonstrating semantic agreement between CLAP-general and retrieved audio. More examples are listed in Table~\ref{tab:clap-eg}.

\begin{table*}[!htbp]
\centering
\fontsize{7pt}{9.75pt}\selectfont
\begin{tabular}{p{2cm} p{2cm} p{2cm} p{2cm} p{2.5cm}}
\toprule
 & \multicolumn{4}{c}{Captions of Retrieved Audios} \\ 
Query Text & Rank 1 & Rank 2 & Rank 3 & Ground Truth (Rank 9) \\ \midrule
\multirow{5}{2cm}{Water flows and people speak in the distance} & Water splashing with multiple voices in background & Water is trickling, and a man talks & A river stream flowing followed by a kid talking & Running water and distant speech \\ \cmidrule{2-5}
 & A man shouting as a stream of water splashes and a crowd of people talk in the background & Splashing water and quiet murmuring & A large volume of water is rushing, splashing and gurgling, and an adult male speaks briefly & A stream of water rushing as a man shouts in the distance \\ \cmidrule{2-5}
 & A plastic clack followed by a man talking as a stream of water rushes and a crowd of people talk in the background & Bubbles gurgling and water spraying as a man speaks softly while crowd of people talk in the background & A stream of water rushing and trickling followed by a young man whooshing & Water rushing loudly while a man yells in the background \\ \cmidrule{2-5}
 & Water splashes and a man speaks & Water trickling and faint, muffled speech & Sounds of a river with man briefly mumbling & A large volume of water is rushing fast, splashing and roaring, and an adult male shout in the background \\ \cmidrule{2-5}
 & Water is falling, splashing and gurgling, a crowd of people talk in the background, and an adult male speaks in the foreground & Water spraying and gurgling as a man speaks and a crowd of people talk in the background & A stream burbles while a man speaks & Water flows and people speak in the distance \\ \midrule
Query Text & Rank 1 & Rank 2 & Rank 3 & Ground Truth (Rank 10) \\ \midrule
\multirow{5}{2cm}{A frog croaks with speech and thumping noises in the background} & Frogs croaking together with a man speaking followed by rustling & Frogs croaking and a humming with insects vocalizing & Frogs croaking with rustling in the background & Nature sounds with a frog croaking \\ \cmidrule{2-5}
 & A man talking followed by plastic clunking and rattling as frogs croak and crickets chirp & A frog croaking and insects vocalizing with a humming & Two instances of bird wings flapping while frogs are croaking & A frog chirping as a woman talks over an intercom and water splashes in the background followed by wood falling on a hard surface \\ \cmidrule{2-5}
 & A man talking followed by plastic creaking and clacking as frogs croak and crickets chirp & A croaking frog with brief bird chirps & A group of frogs croaking as plastic flutters in the background & A frog chirping with distant speaking of a person \\ \cmidrule{2-5}
 & Several frogs chirping near and far with men speaking and some banging & Crickets chirping very loudly & Frogs chirp loudly & A frog croaking as a woman talks through an intercom while water is splashing and wood clanks in the background \\ \cmidrule{2-5}
 & A man speaking as frogs croak and crickets chirp while a motorboat engine runs alongside several plastic clacks and clanging & Ambient horror music plays as birds chirp and frogs croak & High pitched croaking of frogs with some rustling & A frog croaks with speech and thumping noises in the background \\ 
\bottomrule
\end{tabular}
\caption{Captions of Audio retrieved for a Query text (Text-to-Audio retrieval task) using \metal-aligned CLAP-General $\times$ M-MPNET}
\label{tab:clap-eg}
\end{table*}

\section{Cross-lingual Text-to-Image Generation.}
\label{app:text-to-image}
\begin{figure*}[!htbp]
    \centering
    \begin{subfigure}{0.157\textwidth}
        \includegraphics[width=\linewidth]{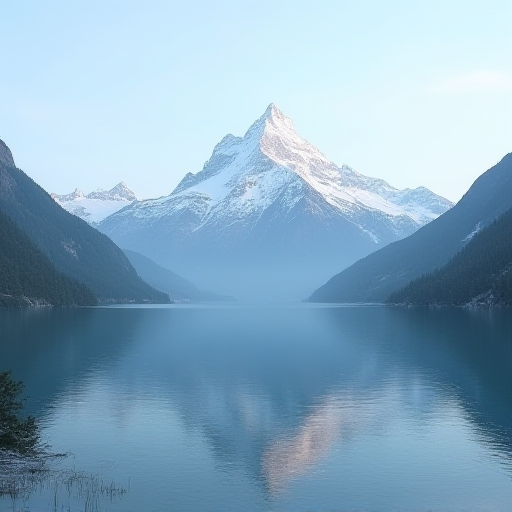}
        \caption{FLUX (en)}
    \end{subfigure}
    \hspace*{\fill}
    \begin{subfigure}{0.157\textwidth}
        \includegraphics[width=\linewidth]{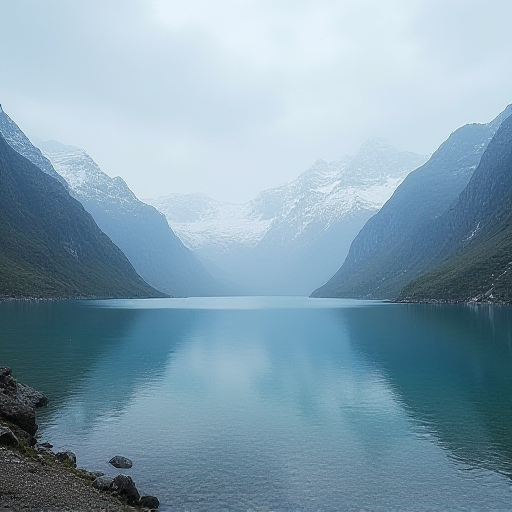}
        \caption{Ours (en)}
    \end{subfigure}
    \hspace*{\fill}
    \begin{subfigure}{0.157\textwidth}
        \includegraphics[width=\linewidth]{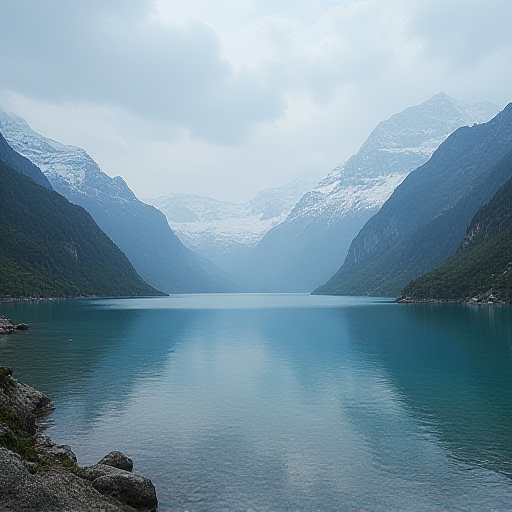}
        \caption{Ours (el)}
    \end{subfigure}
    \hspace*{\fill}
    \begin{subfigure}{0.157\textwidth}
        \includegraphics[width=\linewidth]{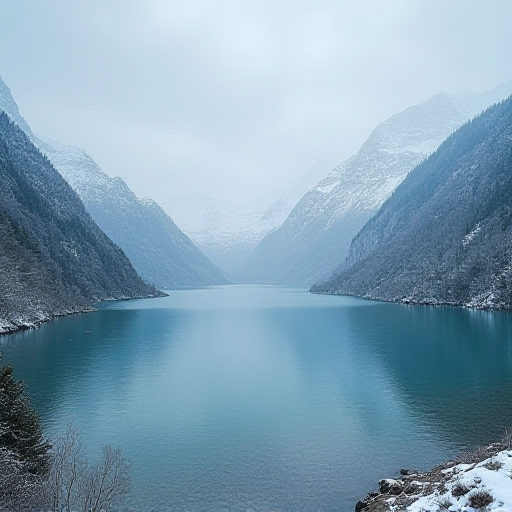}
        \caption{Ours (fa)}
    \end{subfigure}
    \hspace*{\fill}
    \begin{subfigure}{0.157\textwidth}
        \includegraphics[width=\linewidth]{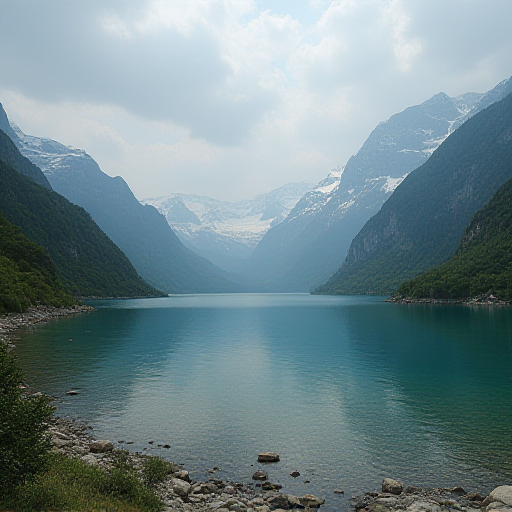}
        \caption{Ours (fr)}
    \end{subfigure}
    \hspace*{\fill}
    \begin{subfigure}{0.157\textwidth}
        \includegraphics[width=\linewidth]{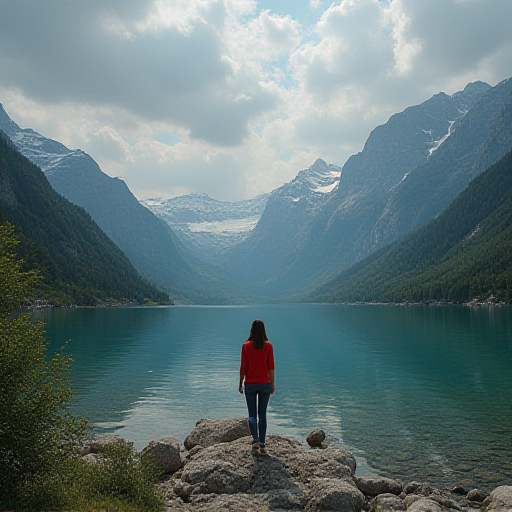}
        \caption{Ours (he)}
    \end{subfigure}

    \vspace{10pt} 

    \begin{subfigure}{0.157\textwidth}
        \includegraphics[width=\linewidth]{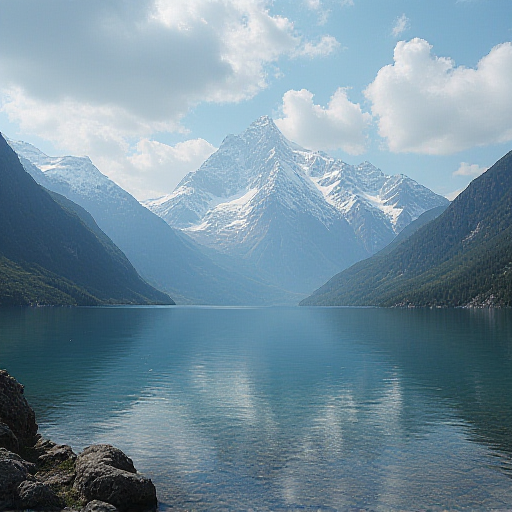}
        \caption{FLUX CLIP (en)}
    \end{subfigure}
    \hspace*{\fill}
    \begin{subfigure}{0.157\textwidth}
        \includegraphics[width=\linewidth]{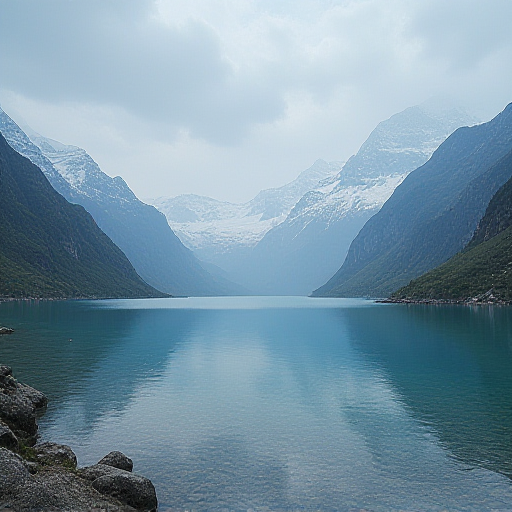}
        \caption{Ours (ru)}
    \end{subfigure}
    \hspace*{\fill}
    \begin{subfigure}{0.157\textwidth}
        \includegraphics[width=\linewidth]{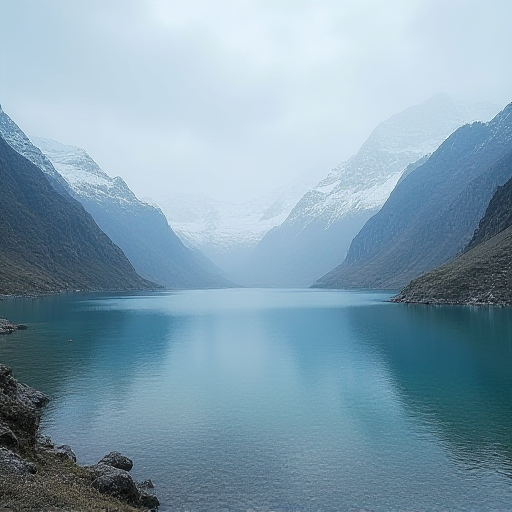}
        \caption{Ours (hi)}
    \end{subfigure}
    \hspace*{\fill}
    \begin{subfigure}{0.157\textwidth}
        \includegraphics[width=\linewidth]{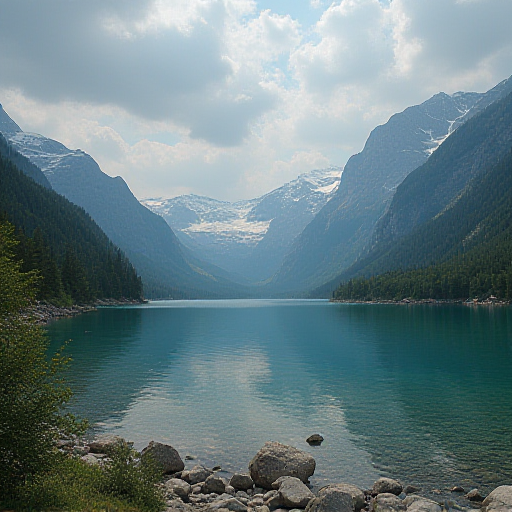}
        \caption{Ours (id)}
    \end{subfigure}
    \hspace*{\fill}
    \begin{subfigure}{0.157\textwidth}
        \includegraphics[width=\linewidth]{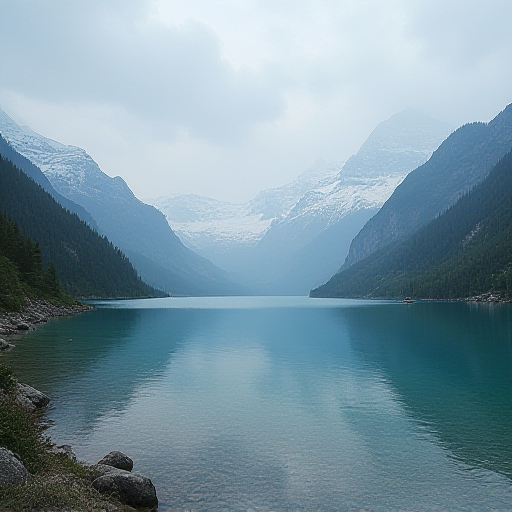}
        \caption{Ours (ko)}
    \end{subfigure}
    \hspace*{\fill}
    \begin{subfigure}{0.157\textwidth}
        \includegraphics[width=\linewidth]{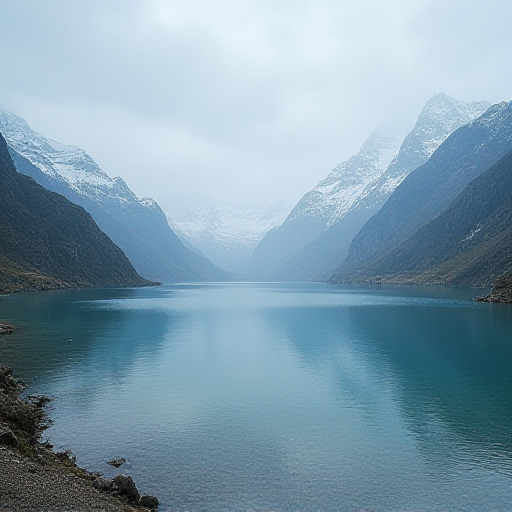}
        \caption{Ours (es)}
    \end{subfigure}

    \caption{Images generated by FLUX text-to-image model using the prompt ``a snow caped mountain is behind a large lake'' in multiple languages. Our \metal-aligned model produces similar quality images compared to baseline FLUX (both T5 and CLIP encoders), and FLUX-CLIP models.}
    \label{fig:flux-snow}
\end{figure*}

\begin{figure*}[!htbp]
    \centering
    \begin{subfigure}{0.157\textwidth}
        \includegraphics[width=\linewidth]{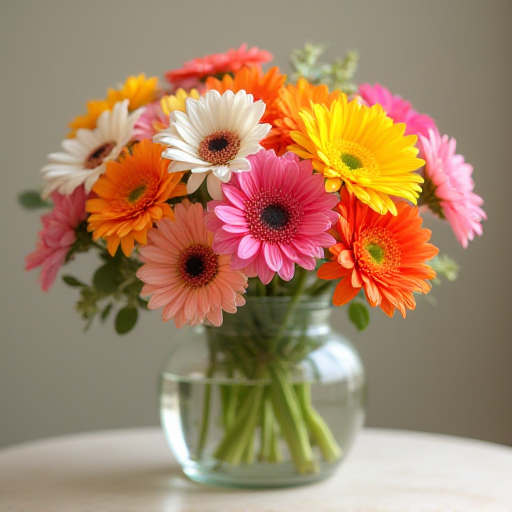}
        \caption{FLUX (en)}
    \end{subfigure}
    \hspace*{\fill}
    \begin{subfigure}{0.157\textwidth}
        \includegraphics[width=\linewidth]{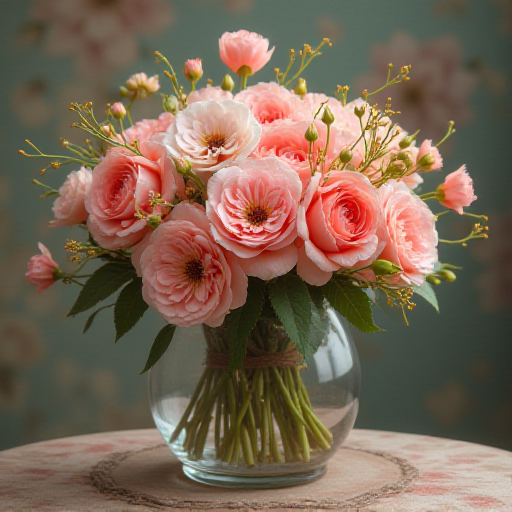}
        \caption{Ours (en)}
    \end{subfigure}
    \hspace*{\fill}
    \begin{subfigure}{0.157\textwidth}
        \includegraphics[width=\linewidth]{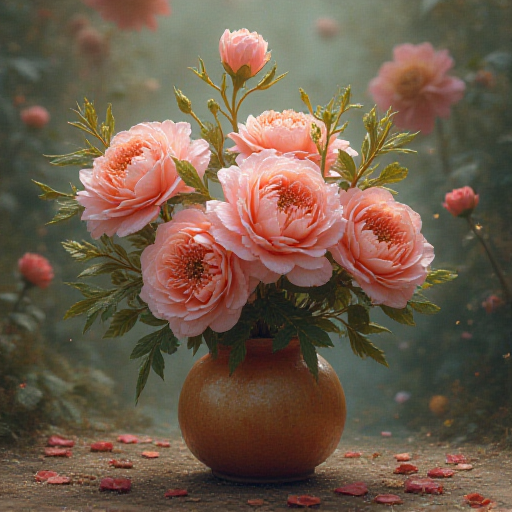}
        \caption{Ours (el)}
    \end{subfigure}
    \hspace*{\fill}
    \begin{subfigure}{0.157\textwidth}
        \includegraphics[width=\linewidth]{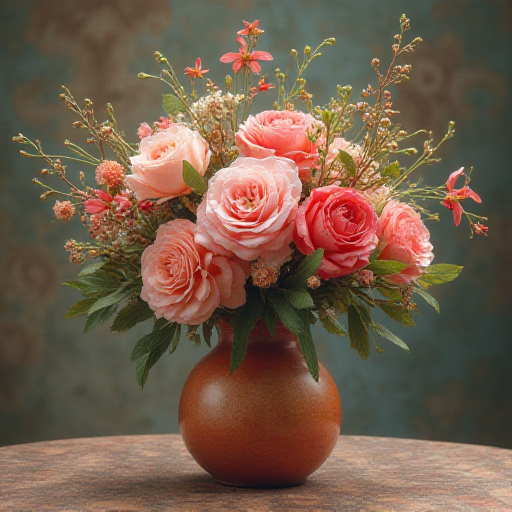}
        \caption{Ours (fa)}
    \end{subfigure}
    \hspace*{\fill}
    \begin{subfigure}{0.157\textwidth}
        \includegraphics[width=\linewidth]{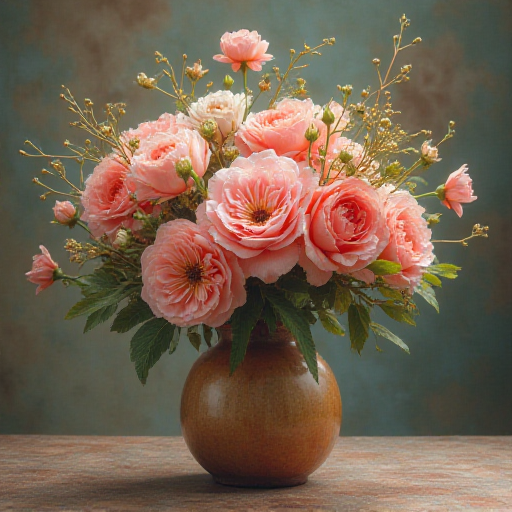}
        \caption{Ours (fr)}
    \end{subfigure}
    \hspace*{\fill}
    \begin{subfigure}{0.157\textwidth}
        \includegraphics[width=\linewidth]{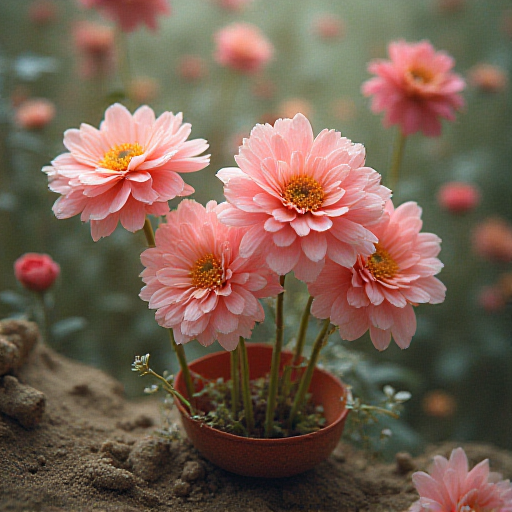}
        \caption{Ours (he)}
    \end{subfigure}

    \vspace{10pt} 

    \begin{subfigure}{0.157\textwidth}
        \includegraphics[width=\linewidth]{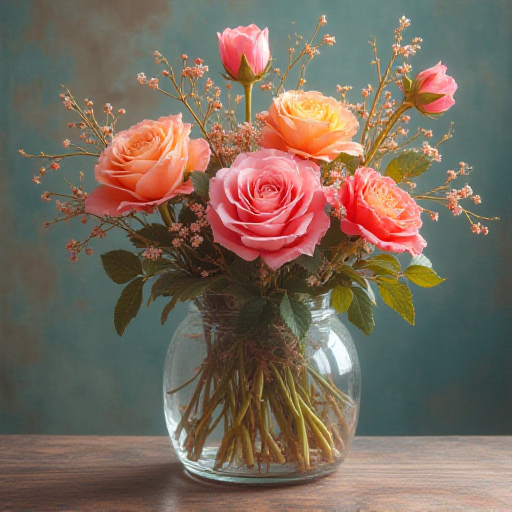}
        \caption{FLUX CLIP (en)}
    \end{subfigure}
    \hspace*{\fill}
    \begin{subfigure}{0.157\textwidth}
        \includegraphics[width=\linewidth]{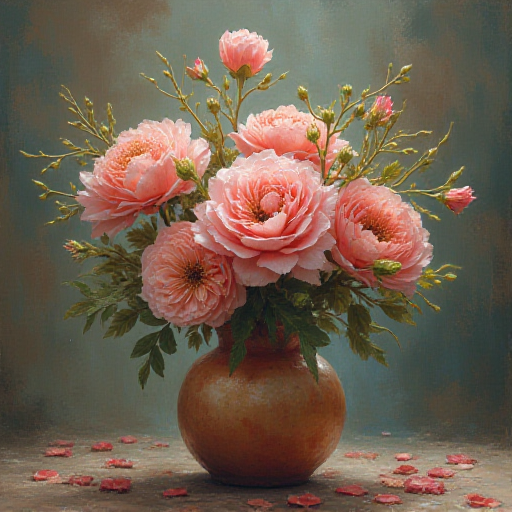}
        \caption{Ours (ru)}
    \end{subfigure}
    \hspace*{\fill}
    \begin{subfigure}{0.157\textwidth}
        \includegraphics[width=\linewidth]{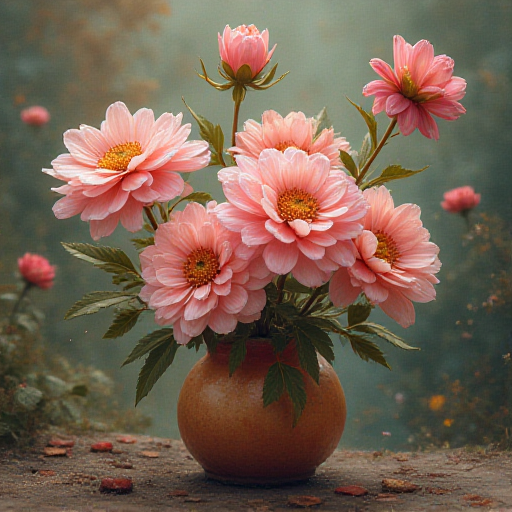}
        \caption{Ours (hi)}
    \end{subfigure}
    \hspace*{\fill}
    \begin{subfigure}{0.157\textwidth}
        \includegraphics[width=\linewidth]{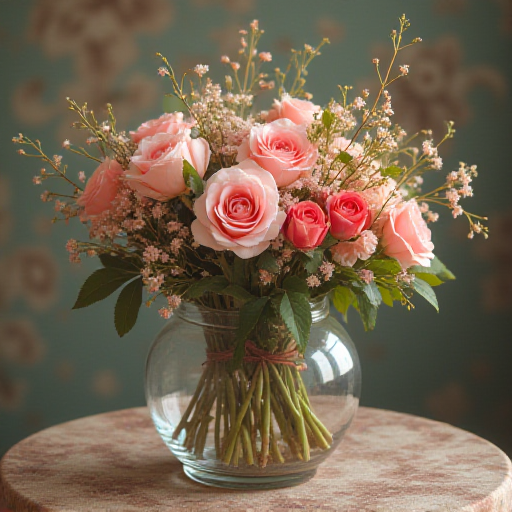}
        \caption{Ours (id)}
    \end{subfigure}
    \hspace*{\fill}
    \begin{subfigure}{0.157\textwidth}
        \includegraphics[width=\linewidth]{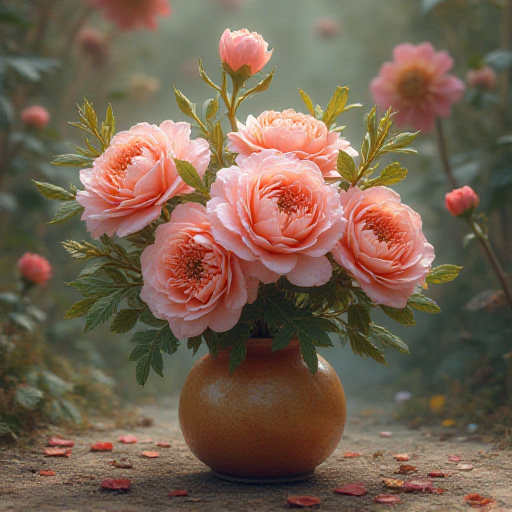}
        \caption{Ours (ko)}
    \end{subfigure}
    \hspace*{\fill}
    \begin{subfigure}{0.157\textwidth}
        \includegraphics[width=\linewidth]{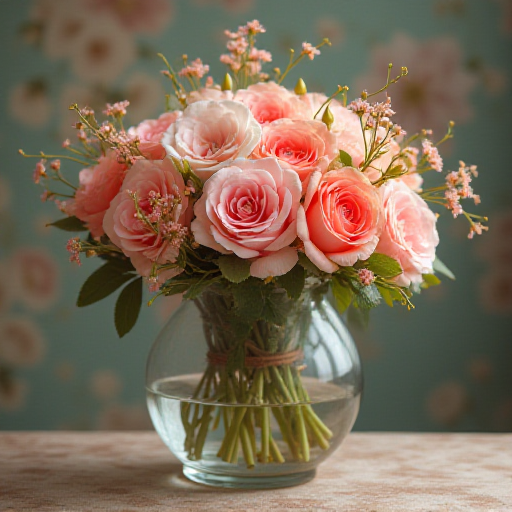}
        \caption{Ours (es)}
    \end{subfigure}

    \caption{Images generated by FLUX text-to-image model using the prompt ``Assortment of colorful flowers in glass vase on table.'' in multiple languages. Our \metal-aligned model produces similar quality images compared to baseline FLUX (both T5 and CLIP encoders), and FLUX-CLIP models.}
    \label{fig:flux-flowers}
\end{figure*}

\begin{figure*}[!htbp]
    \centering
    \captionsetup[subfigure]{labelformat=empty} 

    \begin{minipage}{\textwidth}
        \centering
        {\normalfont (1) T5 prompt: ``A photo of: ''}
        \vspace{10pt}
    \end{minipage}
    
    \begin{subfigure}{0.157\textwidth}
        \includegraphics[width=\linewidth]{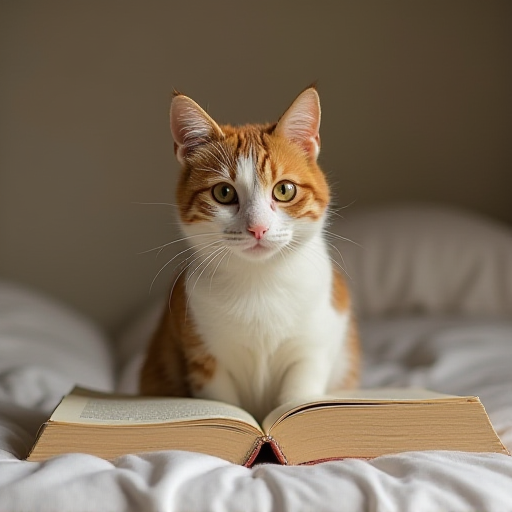}
        \caption{FLUX (en)}
    \end{subfigure}
    \hspace*{\fill}
    \begin{subfigure}{0.157\textwidth}
        \includegraphics[width=\linewidth]{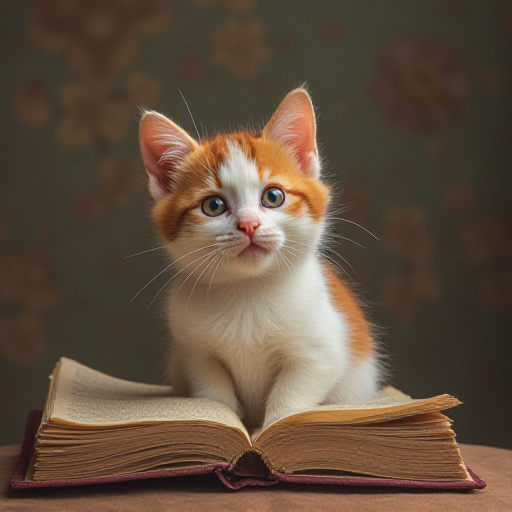}
        \caption{Ours (en)}
    \end{subfigure}
    \hspace*{\fill}
    \begin{subfigure}{0.157\textwidth}
        \includegraphics[width=\linewidth]{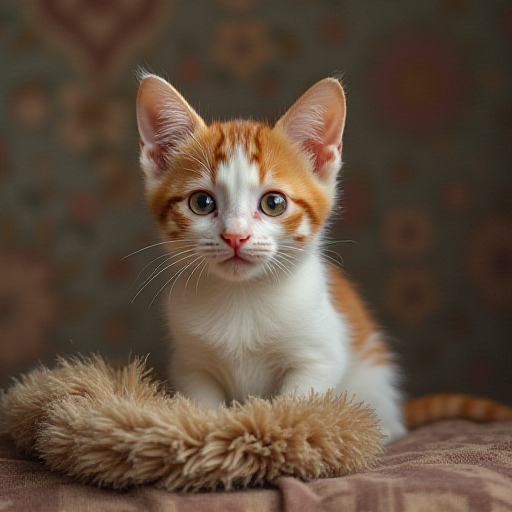}
        \caption{Ours (el)}
    \end{subfigure}
    \hspace*{\fill}
    \begin{subfigure}{0.157\textwidth}
        \includegraphics[width=\linewidth]{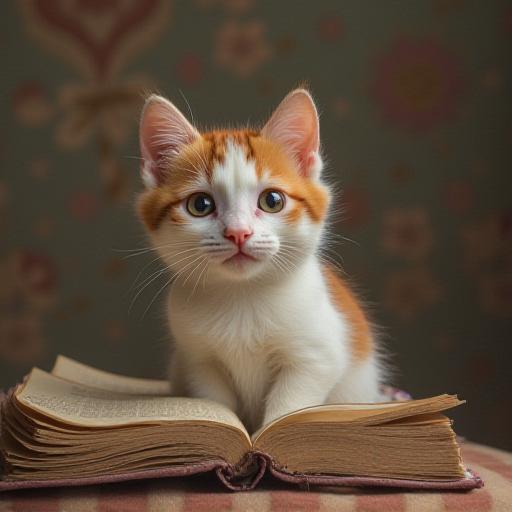}
        \caption{Ours (fa)}
    \end{subfigure}
    \hspace*{\fill}
    \begin{subfigure}{0.157\textwidth}
        \includegraphics[width=\linewidth]{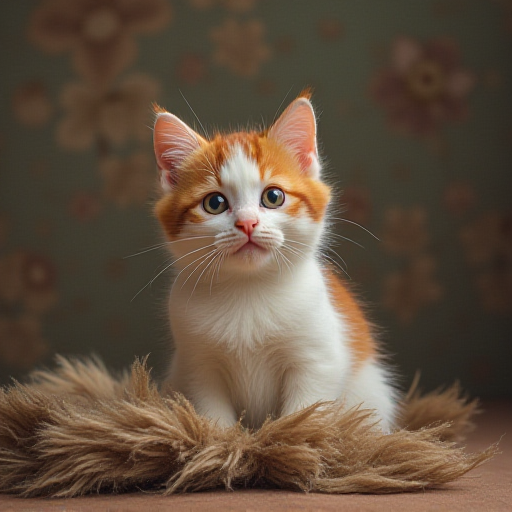}
        \caption{Ours (fr)}
    \end{subfigure}
    \hspace*{\fill}
    \begin{subfigure}{0.157\textwidth}
        \includegraphics[width=\linewidth]{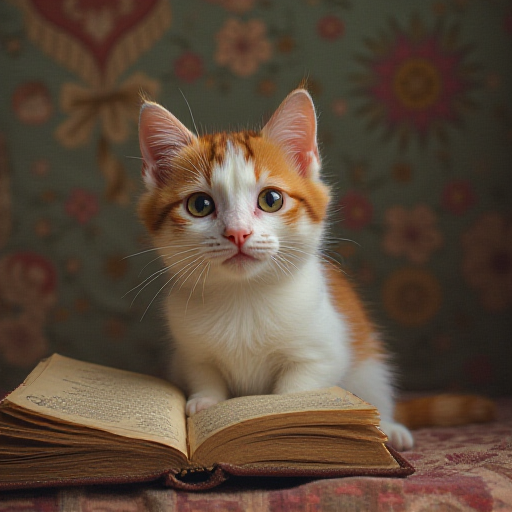}
        \caption{Ours (he)}
    \end{subfigure}

    \vspace{10pt} 

    \begin{subfigure}{0.157\textwidth}
        \includegraphics[width=\linewidth]{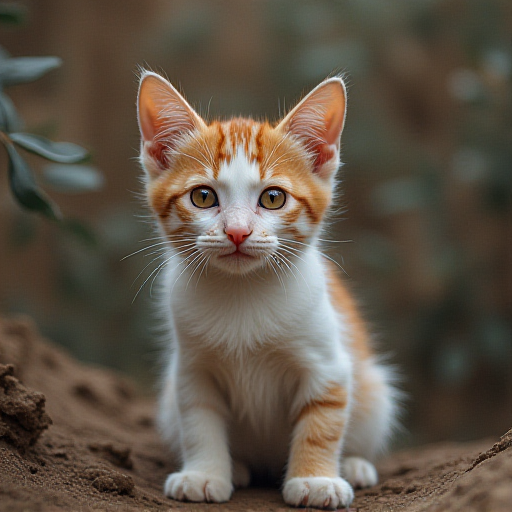}
        \caption{FLUX CLIP (en)}
    \end{subfigure}
    \hspace*{\fill}
    \begin{subfigure}{0.157\textwidth}
        \includegraphics[width=\linewidth]{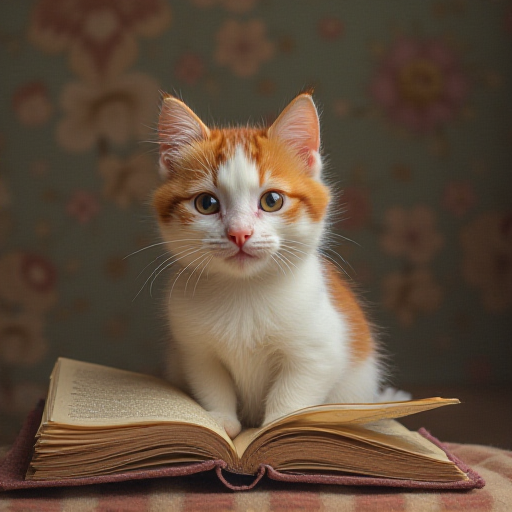}
        \caption{Ours (ru)}
    \end{subfigure}
    \hspace*{\fill}
    \begin{subfigure}{0.157\textwidth}
        \includegraphics[width=\linewidth]{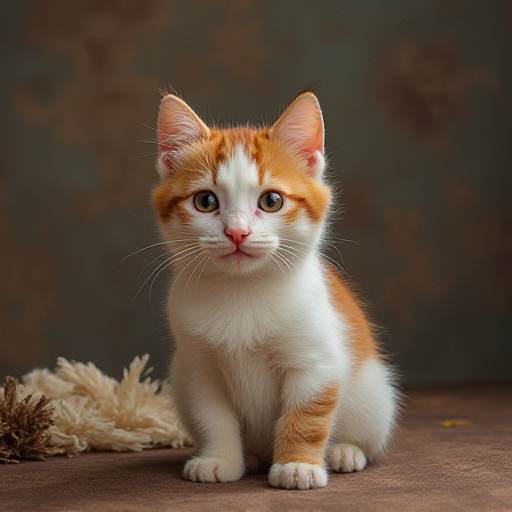}
        \caption{Ours (hi)}
    \end{subfigure}
    \hspace*{\fill}
    \begin{subfigure}{0.157\textwidth}
        \includegraphics[width=\linewidth]{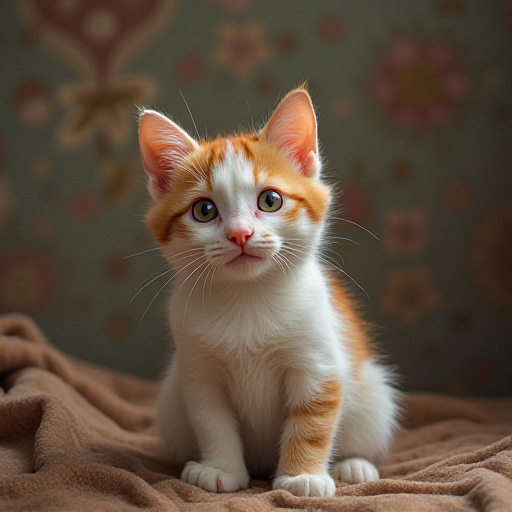}
        \caption{Ours (id)}
    \end{subfigure}
    \hspace*{\fill}
    \begin{subfigure}{0.157\textwidth}
        \includegraphics[width=\linewidth]{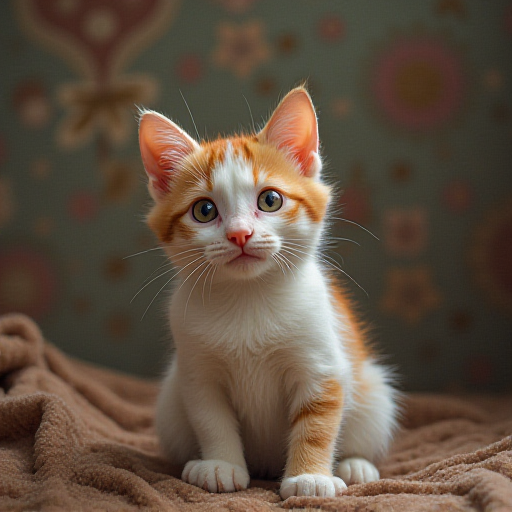}
        \caption{Ours (ko)}
    \end{subfigure}
    \hspace*{\fill}
    \begin{subfigure}{0.157\textwidth}
        \includegraphics[width=\linewidth]{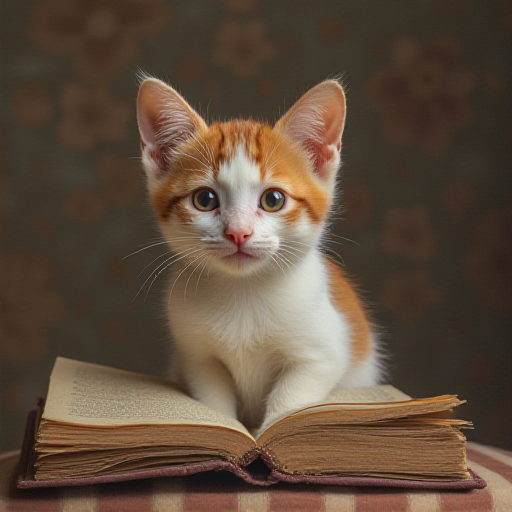}
        \caption{Ours (es)}
    \end{subfigure}

    \begin{minipage}{\textwidth}
        \vspace{10pt}
        \centering
        {\normalfont (2) T5 prompt: ``add a book: ''}
        \vspace{10pt}
    \end{minipage}
    
    \begin{subfigure}{0.157\textwidth}
        \includegraphics[width=\linewidth]{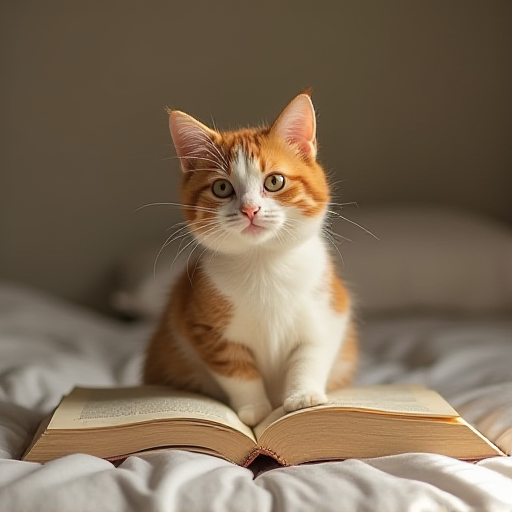}
        \caption{FLUX-T5 (en)}
    \end{subfigure}
    \hspace*{\fill}
    \begin{subfigure}{0.157\textwidth}
        \includegraphics[width=\linewidth]{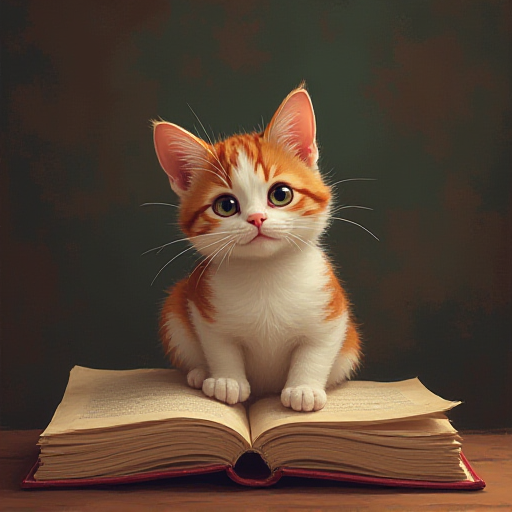}
        \caption{Ours (en)}
    \end{subfigure}
    \hspace*{\fill}
    \begin{subfigure}{0.157\textwidth}
        \includegraphics[width=\linewidth]{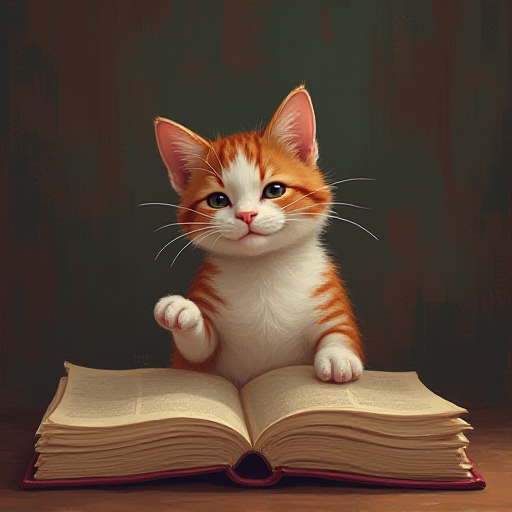}
        \caption{Ours (el)}
    \end{subfigure}
    \hspace*{\fill}
    \begin{subfigure}{0.157\textwidth}
        \includegraphics[width=\linewidth]{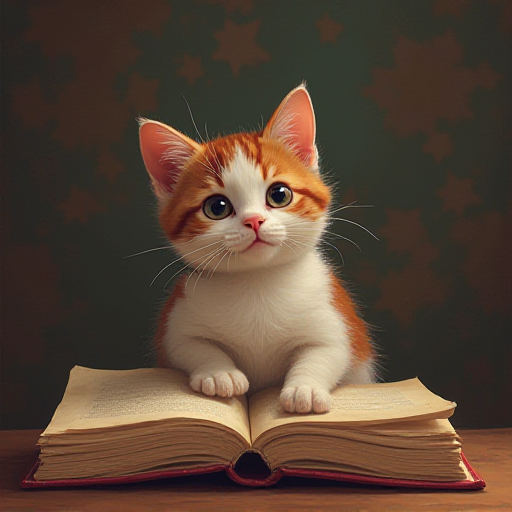}
        \caption{Ours (fa)}
    \end{subfigure}
    \hspace*{\fill}
    \begin{subfigure}{0.157\textwidth}
        \includegraphics[width=\linewidth]{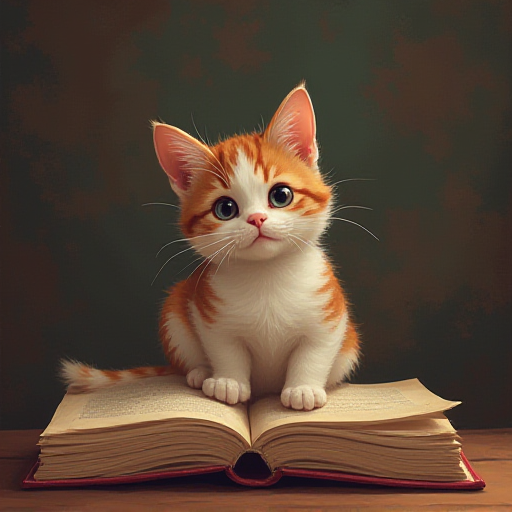}
        \caption{Ours (fr)}
    \end{subfigure}
    \hspace*{\fill}
    \begin{subfigure}{0.157\textwidth}
        \includegraphics[width=\linewidth]{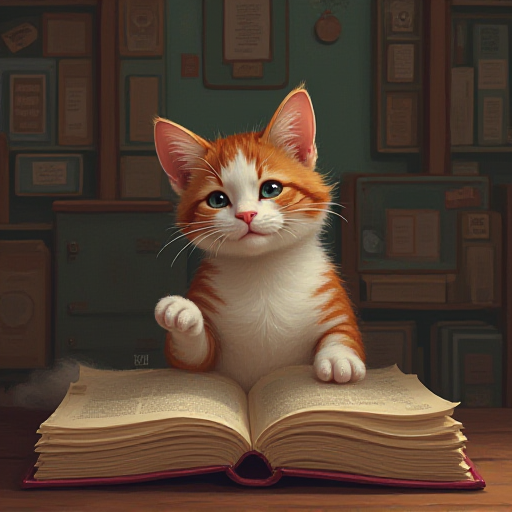}
        \caption{Ours (he)}
    \end{subfigure}

    \vspace{10pt} 

    \begin{subfigure}{0.157\textwidth}
        \includegraphics[width=\linewidth]{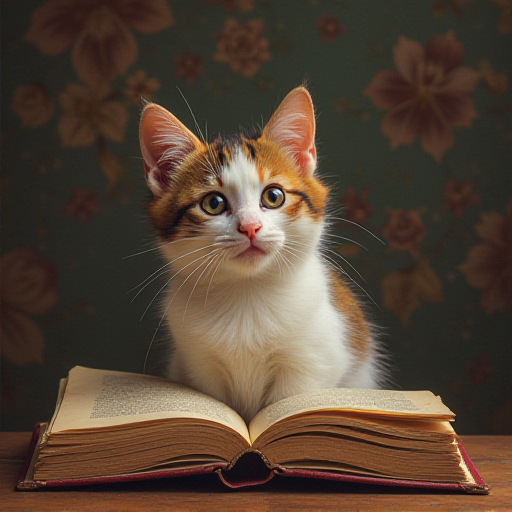}
        \caption{FLUX CLIP (en)}
    \end{subfigure}
    \hspace*{\fill}
    \begin{subfigure}{0.157\textwidth}
        \includegraphics[width=\linewidth]{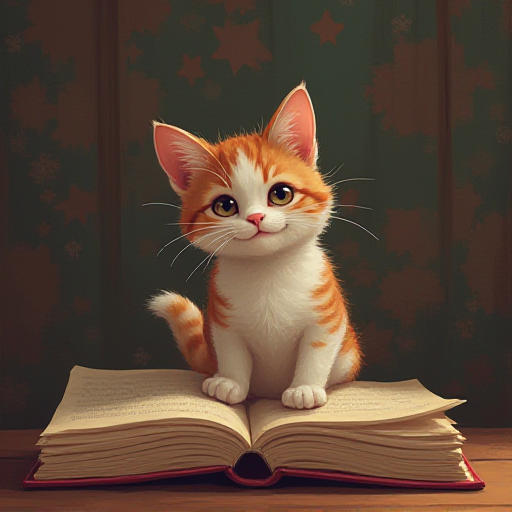}
        \caption{Ours (ru)}
    \end{subfigure}
    \hspace*{\fill}
    \begin{subfigure}{0.157\textwidth}
        \includegraphics[width=\linewidth]{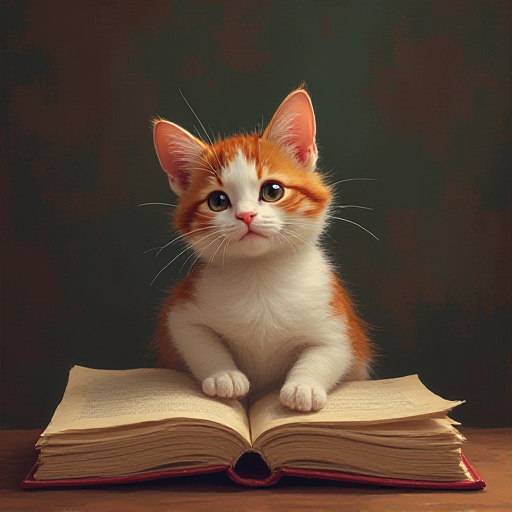}
        \caption{Ours (hi)}
    \end{subfigure}
    \hspace*{\fill}
    \begin{subfigure}{0.157\textwidth}
        \includegraphics[width=\linewidth]{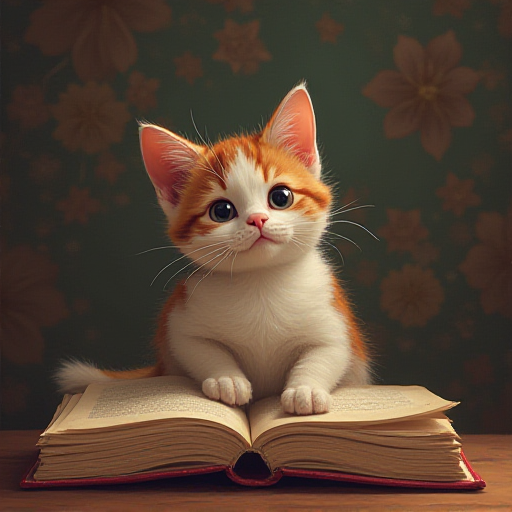}
        \caption{Ours (id)}
    \end{subfigure}
    \hspace*{\fill}
    \begin{subfigure}{0.157\textwidth}
        \includegraphics[width=\linewidth]{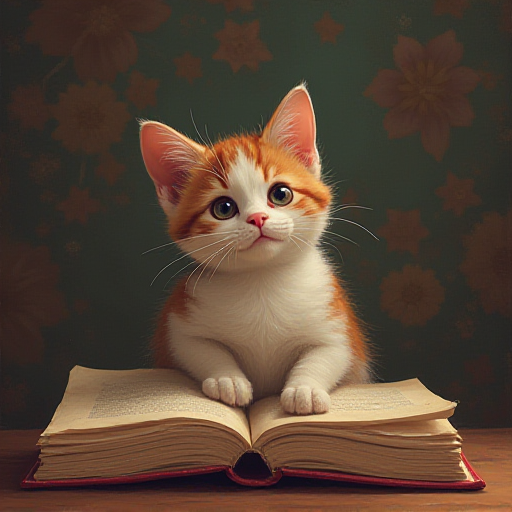}
        \caption{Ours (ko)}
    \end{subfigure}
    \hspace*{\fill}
    \begin{subfigure}{0.157\textwidth}
        \includegraphics[width=\linewidth]{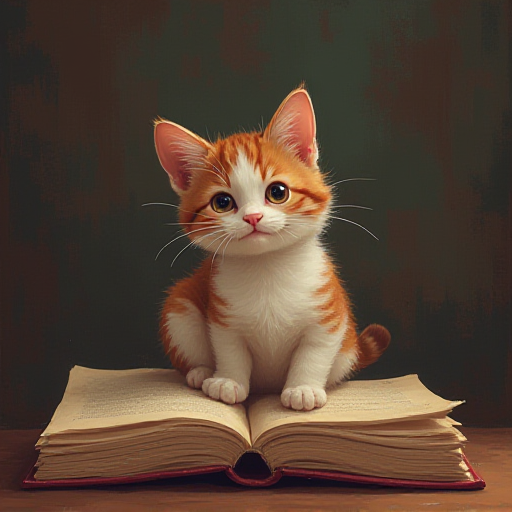}
        \caption{Ours (es)}
    \end{subfigure}

    \begin{minipage}{\textwidth}
        \centering
        \vspace{10pt}
        {\normalfont (3) T5 prompt: ``add a book on bed: ''}
        \vspace{10pt}
    \end{minipage}
    
    \begin{subfigure}{0.157\textwidth}
        \includegraphics[width=\linewidth]{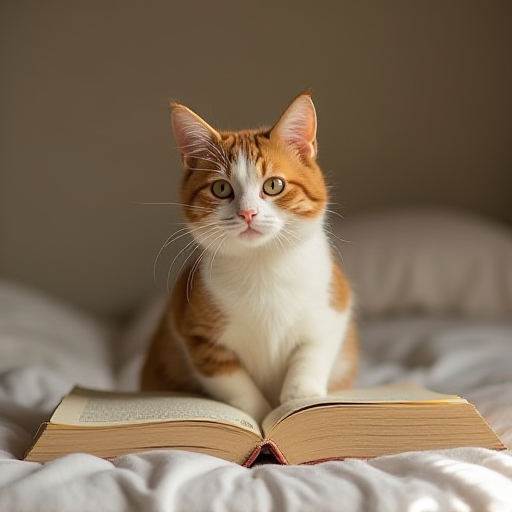}
        \caption{FLUX-T5 (en)}
    \end{subfigure}
    \hspace*{\fill}
    \begin{subfigure}{0.157\textwidth}
        \includegraphics[width=\linewidth]{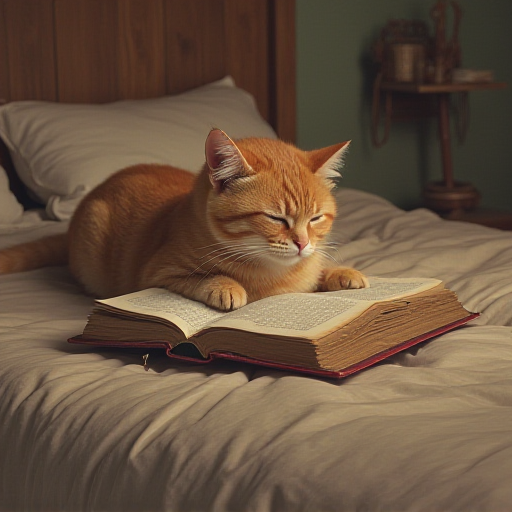}
        \caption{Ours (en)}
    \end{subfigure}
    \hspace*{\fill}
    \begin{subfigure}{0.157\textwidth}
        \includegraphics[width=\linewidth]{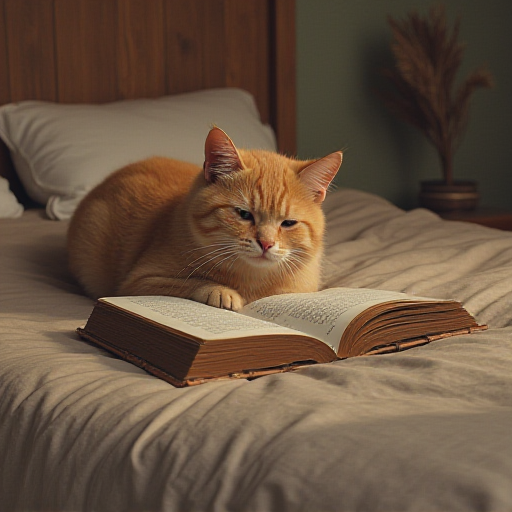}
        \caption{Ours (el)}
    \end{subfigure}
    \hspace*{\fill}
    \begin{subfigure}{0.157\textwidth}
        \includegraphics[width=\linewidth]{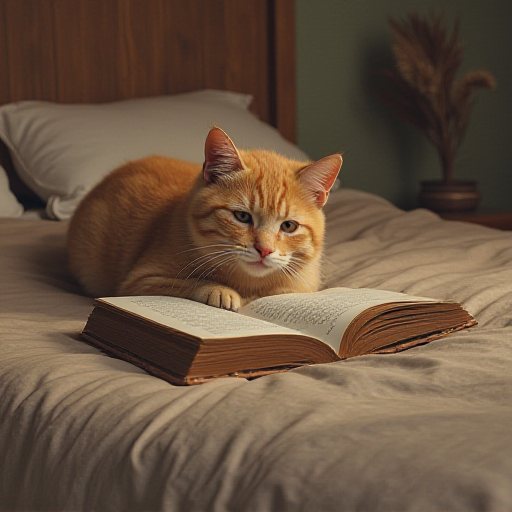}
        \caption{Ours (fa)}
    \end{subfigure}
    \hspace*{\fill}
    \begin{subfigure}{0.157\textwidth}
        \includegraphics[width=\linewidth]{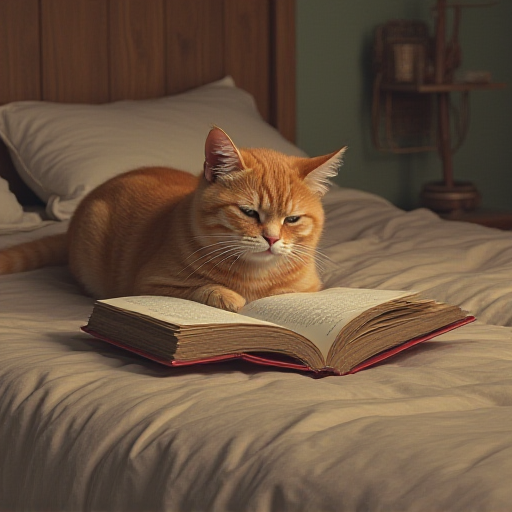}
        \caption{Ours (fr)}
    \end{subfigure}
    \hspace*{\fill}
    \begin{subfigure}{0.157\textwidth}
        \includegraphics[width=\linewidth]{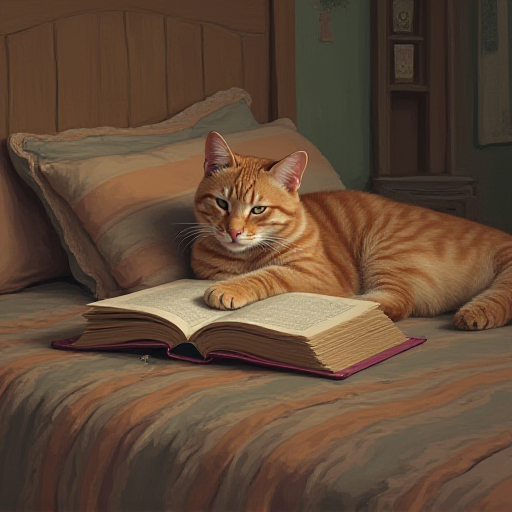}
        \caption{Ours (he)}
    \end{subfigure}

    \vspace{10pt} 

    \begin{subfigure}{0.157\textwidth}
        \includegraphics[width=\linewidth]{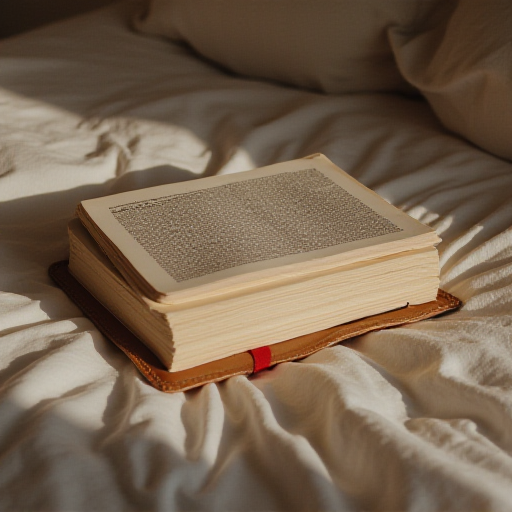}
        \caption{FLUX CLIP (en)}
    \end{subfigure}
    \hspace*{\fill}
    \begin{subfigure}{0.157\textwidth}
        \includegraphics[width=\linewidth]{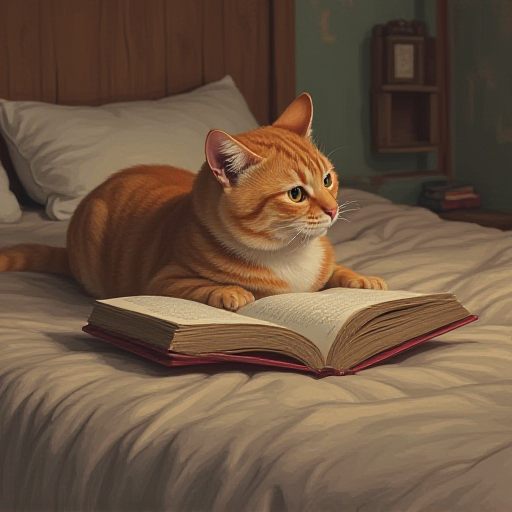}
        \caption{Ours (ru)}
    \end{subfigure}
    \hspace*{\fill}
    \begin{subfigure}{0.157\textwidth}
        \includegraphics[width=\linewidth]{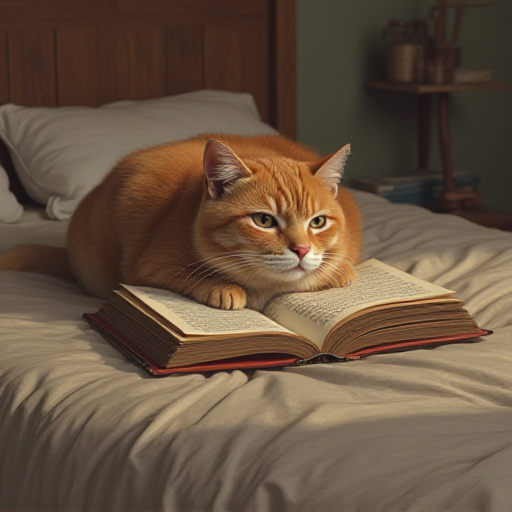}
        \caption{Ours (hi)}
    \end{subfigure}
    \hspace*{\fill}
    \begin{subfigure}{0.157\textwidth}
        \includegraphics[width=\linewidth]{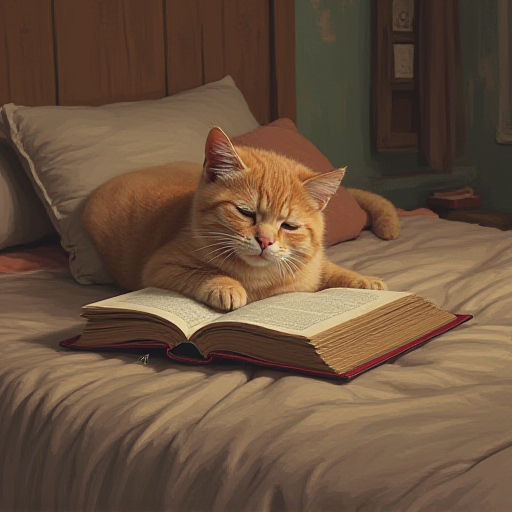}
        \caption{Ours (id)}
    \end{subfigure}
    \hspace*{\fill}
    \begin{subfigure}{0.157\textwidth}
        \includegraphics[width=\linewidth]{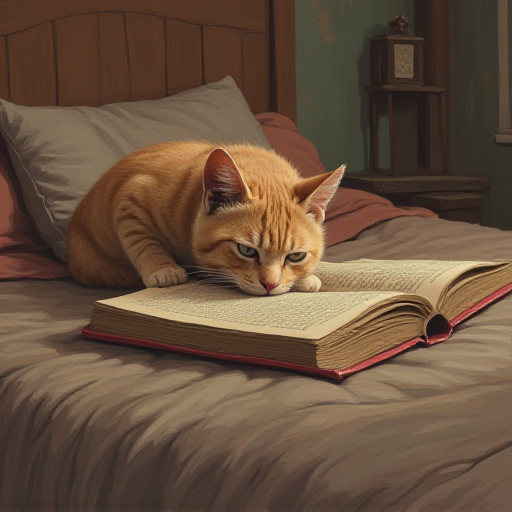}
        \caption{Ours (ko)}
    \end{subfigure}
    \hspace*{\fill}
    \begin{subfigure}{0.157\textwidth}
        \includegraphics[width=\linewidth]{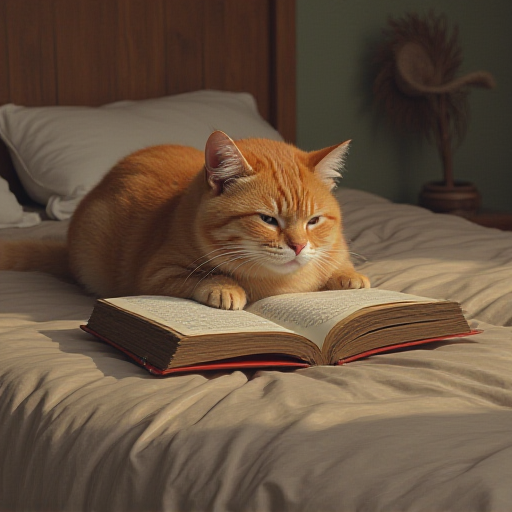}
        \caption{Ours (es)}
    \end{subfigure}

    \caption{Images generated by FLUX models using the prompt ``A cat sitting on a bed behind a book'' in multiple languages. Our \metal-aligned model produces similar images but with missing objects (book, bed) compared to FLUX models (T5 and CLIP encoders). T5 prompts help mitigate this issue, as shown in sub-figures (2) \& (3).}
    \label{fig:flux-cat}
\end{figure*}

Both Inception score and FID scores are computed using torch-fidelity~\cite{obukhov2020torchfidelity} package~\footnote{\url{https://github.com/toshas/torch-fidelity}}. Language-wise FID scores shown in Table~\ref{tab:fid} and inception scores are shown in Table~\ref{tab:IS}. For English, our aligned model gives better FID score than FLUX-CLIP though both are still high compared to FLUX (upper-bound/skyline model). More examples of generated images are shown in Figure~\ref{fig:flux-snow}, Figure~\ref{fig:flux-flowers} \& Figure~\ref{fig:flux-cat}.
\begin{table*}[!htbp]
\centering
\fontsize{7pt}{9.75pt}\selectfont
\begin{tabular}{l M{0.9cm} M{0.9cm} M{0.9cm} M{0.9cm} M{0.9cm} M{0.9cm} M{0.9cm} M{0.9cm} M{0.9cm} M{0.9cm}}
\toprule
\multirow{2}{*}{Models} & \multicolumn{10}{c}{Inception Score ($\uparrow$)} \\
 & en & fr & el & he & id & ko & fa & ru & es & hi \\ \midrule
FLUX & \plusminus{42.3}{0.81} & - & - & - & - & - & - & - & - & - \\
FLUX-T5 & \plusminus{42.1}{0.64} & - & - & - & - & - & - & - & - & - \\
FLUX-CLIP & \plusminus{33.4}{0.57} & - & - & - & - & - & - & - & - & - \\
FLUX $\times$ M-MPNET & \plusminus{35.9}{0.57} & \plusminus{32.7}{0.80} & \plusminus{29.9}{0.66} & \plusminus{29.9}{0.45} & \plusminus{34.3}{0.76} & \plusminus{30.2}{0.51} & \plusminus{32.5}{0.74} & \plusminus{28.6}{0.63} & \plusminus{32.8}{0.50} & \plusminus{31.3}{0.46} \\

\bottomrule
\end{tabular}
\caption{Inception score for MSCOCO-30K on 512$\times$512 images (10 inference steps; guidance scale = 3.5).}
\label{tab:IS}
\end{table*}
\begin{table*}[!htbp]
\centering
\fontsize{7pt}{9.75pt}\selectfont
\begin{tabular}{l c c c c c c c c c c}
\toprule
\multirow{2}{*}{Models} & \multicolumn{10}{c}{FID-30K ($\downarrow$)} \\
 & en & fr & el & he & id & ko & fa & ru & es & hi \\ \midrule
FLUX & 23.4 & - & - & - & - & - & - & - & - & - \\
FLUX-T5 & 23.4 & - & - & - & - & - & - & - & - & - \\
FLUX-CLIP & 40.9 & - & - & - & - & - & - & - & - & - \\
FLUX $\times$ M-MPNET & 36.9 & 41.8 & 46.6 & 46.9 & 40.0 & 45.4 & 43.0 & 47.2 & 41.1 & 45.1 \\
\bottomrule
\end{tabular}
\caption{FID scores computed on our MSCOCO 30K synthetic multilingual evaluation dataset.}
\label{tab:fid}
\end{table*}





\end{document}